%% file: main_arxiv.tex
\documentclass[11pt]{article}

\usepackage[margin=1in]{geometry}
\usepackage{mathpazo}
\usepackage[backend=biber,style=alphabetic,natbib=true,maxnames=99,maxcitenames=2,minalphanames=3]{biblatex}
\input{header}

\title{Is a Caption Worth a Thousand Images? \\
      A Controlled Study for Representation Learning}
\author{
    Shibani Santurkar \\
  Stanford \\
  \texttt{shibani@stanford.edu} \\
  \and
  Yann Dubois \\
  Stanford \\
  \texttt{yanndubs@stanford.edu} \\
  \and
  Rohan Taori \\
  Stanford \\
  \texttt{rtaori@stanford.edu} \\
  \and
  Percy Liang \\
  Stanford \\
  \texttt{pliang@cs.stanford.edu} \\
  \and
  Tatsunori Hashimoto \\
  Stanford \\
  \texttt{thashim@stanford.edu} \\
  }
\date{}

\begin{document}

\maketitle

\input{sections/abstract}

\input{sections/introduction}
\input{sections/apples}
\input{sections/quality}
\input{sections/diversity}
\input{sections/prescription}
\input{sections/related}
\input{sections/conclusion}
\input{sections/impact}
\input{sections/acknowledgements}

\medskip

\printbibliography



\clearpage
\appendix

\input{appendix/setup.tex}
\input{appendix/expts.tex}

\end{document}

%% file: header.tex
\usepackage{graphicx}
\usepackage{psfrag}
\usepackage{amsmath}
\usepackage{amsfonts}
\usepackage{verbatim}
\usepackage{mathrsfs}
\usepackage{hyperref}
\usepackage{amssymb}
\usepackage{pifont}

\usepackage{subcaption}
\usepackage{multirow}
\usepackage{algpseudocode}

\usepackage{color, colortbl}
\usepackage{xspace}
\usepackage[textsize=scriptsize]{todonotes}

\newcommand*\zot{\rotatebox{90}}
\newcommand{\cmark}{\ding{51}}%
\newcommand{\xmark}{\ding{55}}%
\newcommand{\clipd}{CLIP$_{\text{S}}$}%
\newcommand{\simclr}{SimCLR }%
\newcommand{\var}{variability}%

\definecolor{LightCyan}{rgb}{0.92,1,1}
\definecolor{LightGreen}{rgb}{0.9,1,0.9}
\definecolor{LightRed}{rgb}{1,0.95,0.95}
\definecolor{Gray}{gray}{0.9}

\newcommand*{\eg}{e.g.\@\xspace}
\newcommand*{\ie}{i.e.\@\xspace}

%% file: sections/abstract.tex
\begin{abstract}
  The development of CLIP~\cite{radford2021learning} has sparked a debate on whether language supervision can result in vision models with {more transferable representations} than traditional image-only methods.
  Our work studies this question through a carefully controlled comparison of two approaches in terms of their ability to learn representations that generalize to downstream classification tasks.
We find that when the pre-training dataset meets certain criteria---it is sufficiently large and contains descriptive captions with low variability---image-only methods \emph{do not} match CLIP's transfer performance, even when they are trained with more image data. 
However, contrary to what one might expect, there are practical settings in which these criteria are not met, wherein added supervision through captions is actually \emph{detrimental}.
Motivated by our findings, we devise simple prescriptions to enable CLIP to better leverage the language information present in existing pre-training datasets.
\end{abstract}

%% file: sections/introduction.tex
\section{Introduction}
\label{sec:intro}

Image-based contrastive learning approaches have shown promise in building models that generalize beyond the data distributions they are trained on~\citep{wu2018unsupervised,he2020momentum,chen2020simple,caron2020unsupervised,chen2020big,chen2020big,caron2021emerging,chen2021exploring}.
By leveraging large-scale (unlabelled) data sources through self-supervised training, these models learn representations that transfer to diverse downstream tasks---more so that their supervised counterparts~\citep{ericsson2021how}.

Recently, \citet{radford2021learning} showed that a different approach---contrastive learning with language supervision---can yield models (known as CLIP) with remarkable transfer capabilities.
This development has garnered significant interest in the vision and natural language processing communities alike, leading to a debate on the utility of multi-modality in representation learning~\cite{zhai2022lit,devillers2021does,fang2022data}.
Our work focuses on a specific question within this debate:

\begin{center}
	\emph{Does language supervision lead to more transferable representations than using images alone?}
\end{center}

It might seem like the answer to this question is obvious.
After all, CLIP utilized caption information unavailable to traditional image-based approaches and showed substantial gains over prior work \cite{radford2021learning}.
However, CLIP is drastically different from these approaches in many ways, from training data to fine-grained implementation choices~\cite{devillers2021does}, which makes it difficult to isolate the contribution of language supervision.
Further, recent studies on CLIP's zero-shot classification and robustness properties cast doubt on whether adding language supervision is always  beneficial~\cite{fang2022data}. 
Resolving the aforementioned debate thus requires a carefully controlled comparison of the two approaches in which the \emph{only} difference is the form of supervision.

\begin{figure}[!t]
	\centering
	\includegraphics[width=0.825\textwidth]{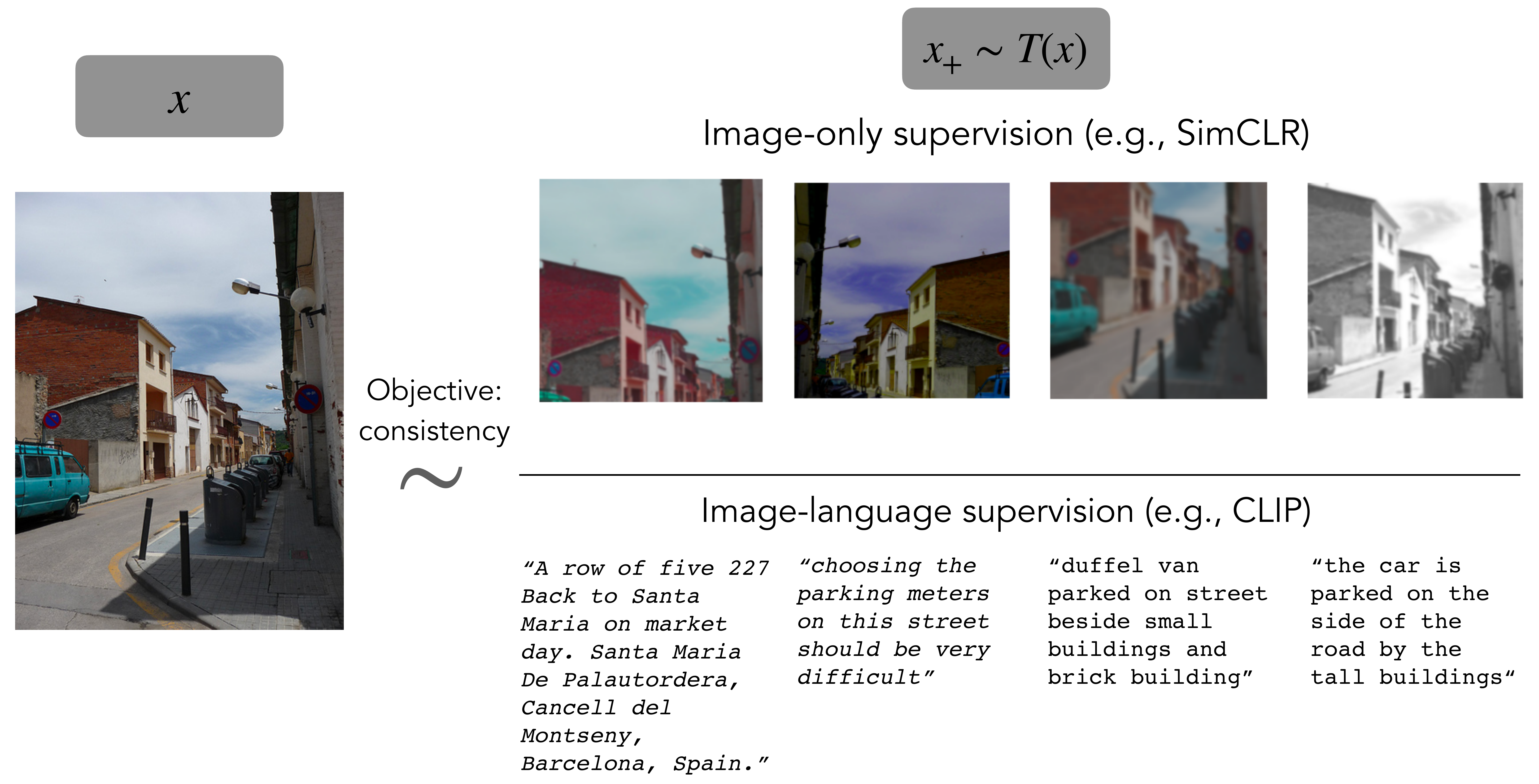}
	\caption{A conceptual view of contrastive image-only and image-language pre-training. The two methods rely on the same self-supervised objective: aligning the representations of 
	positive pairs $(x, x_{+})$ while distinguishing them from negative examples (\eg, other examples in the batch).
	The transformation $T(\cdot)$ which is used to obtain $x_{+}\sim T(x)$ (augmented image or caption) encodes the equivalences we would like the model to satisfy.
  }
	\label{fig:intro_fig}
\end{figure}

\paragraph{Our contributions.}
We devise a methodology to assess the utility of language supervision from a representation learning standpoint.
To do so, we recognize that CLIP and popular 
image-based methods share the same underlying 
primitive of contrastive learning.
In particular, CLIP is conceptually strikingly similar to  
SimCLR~\citep{chen2020simple}.
Perhaps the only irreducible difference between them is whether supervision is provided to the model via image augmentations or image-caption matching (see Figure~\ref{fig:intro_fig})---which is precisely the quantity we want to study.
Thus, we can assess the value of language supervision by systematically comparing appropriately matched versions of \simclr and CLIP\footnote{We use CLIP to mean models trained using \citet{radford2021learning}'s approach, and not their pre-trained model.}  in terms of their downstream transfer performance.
We find that in practice, the picture is nuanced and depends on three properties of the pre-training dataset:

\begin{enumerate}
	\item If the \emph{scale} of the dataset is sufficiently large, CLIP representations transfer better than their \simclr counterparts.
          This gap is not bridged by training \simclr with more data, suggesting that a caption can be worth more than \emph{any} number of images. 
          However, in the low-data regime, language supervision actually hurts model performance in and out-of-distribution.
	\item The \emph{descriptiveness}~\cite{kreiss2021concadia} of dataset captions---\ie, the extent to which they report what is contained in an image---directly determines how well the resulting CLIP models transfer.
          In fact, we find that a single descriptive image-caption pair (e.g., from MS-COCO~\cite{lin2014microsofta}) is worth five less descriptive, uncurated captions (e.g., from YFCC~\cite{thomee2016yfcc100m}).

	\item The \emph{\var{}} of captions within a dataset (\eg due to stylistic or lexical factors) can adversely affect CLIP's  performance.
	We propose a modification to standard CLIP training---performing text data augmentations by sampling from a pool of captions for each image---to alleviate this effect.

\end{enumerate}

Overall, we find that these three properties have inter-twined effects on CLIP's performance.
For instance, the scale of the widely-used YFCC dataset can, to some extent, compensate for its less-descriptive and variable captions.
Guided by our findings, we devise simple interventions on datasets that can lead to more-transferrable CLIP models: (i) filtering out low-quality captions through a text-based classifier, and (ii) applying data augmentation to captions by paraphrasing them using pre-trained language models~\cite{wang2021gpt-j-6b}.

%% file: sections/apples.tex
\section{An apples-to-apples 
comparison}
\label{sec:apples}

As stated in Section~\ref{sec:intro}, our goal is to assess the value of language supervision in representation learning relative to using images alone.
While there have been studies of image-only and image-language pre-training methods in isolation~\cite{wu2018unsupervised,he2020momentum,chen2020simple,caron2020unsupervised,chen2020big,chen2020big,chen2021exploring,caron2021emerging,radford2021learning} and side-by-side~\citep{devillers2021does,fang2022data},
none of these works conclusively answer our motivating question due to confounders such as: (i) algorithmic and architectural variations, and (ii) differing pre-training datasets.

In this section, we outline a series of steps that we take to mitigate these confounders and compare the two methods on equal footing.
Since our focus is on representation learning, we measure performance in terms of the usefulness of a model's representations for downstream tasks, using the evaluation suite of \citet{kornblith2019do} to do so.
We focus on the \emph{fixed-feature} setting where we freeze the weights of a given model and then train a linear probe using task data.
Details of our experimental setup are presented in Appendix~\ref{app:setup}.

\subsection{Finding common ground}
\label{sec:common_ground}
Our approach for isolating the effects of language supervision is guided by the following insight: CLIP shares a fundamental commonality with widely-used image-only pre-training methods.
Namely, they rely on the same algorithmic primitive of \emph{contrastive learning}, which we illustrate in Figure~\ref{fig:intro_fig}.
In both cases, the model is trained with a self-supervised objective: given an image $x$, it must distinguish positive examples $x_+\sim T(x)$ from negative ones (\eg, other examples $\hat{x}$ in the batch).
The choice of transformation $T(\cdot)$ is at the core of contrastive learning as it controls the equivalences encoded in model representations~\cite{dubois2021lossy,haochen2021provable}.
$T(x)$ corresponds to image augmentations (\eg, rotations) in image-only methods and natural language captions in CLIP.

Thus, to understand the role of language supervision, we can compare CLIP to its closest image-only equivalent: \simclr.\footnote{Other image-based methods~\cite{he2020momentum,chen2020big,chen2021exploring,caron2021emerging} have optimizations that are not present in CLIP.}
Both CLIP and SimCLR rely on cross-entropy based objective, which for a given pair $(x, x_+)$ of positive examples with associated negatives $\mathcal{N}$ is
\newcommand{\siml}{\text{sim}}
\begin{eqnarray}
	\ell = - \log \frac{\exp(\siml(z, z_+)/\tau)}{\sum_{n \in \mathcal{N} \cup\{ z_+\} } \exp(\siml(z, z_{n}) / \tau)}\ ,  \ \  \text{where } z = g(\phi(x)) \text{ and } z_{+/n} = g'(\phi'(x_{+/n})) \label{eq:loss},
\end{eqnarray}
where $\text{sim}$ is cosine similarity, $\phi/\phi'$ are encoders, and $g/g'$ are projection heads. 

We now discuss the steps we take to alleviate other inconsistencies (aside from $T(x)$) between CLIP and SimCLR:
\begin{itemize}
\item \emph{Transformation stochasticity:} We first note that the two methods differ in how they obtain $x_+$, not just due to the choice of $T(x)$ but also the generative process itself.
 In \simclr, $x_+$ is a new random draw from $T(x)$ in every batch, while for CLIP, it is a single fixed caption.
Perfectly matching them requires training CLIP by sampling a fresh caption $x_+$ for each image at each iteration. We will refer to this idealized version of CLIP as \clipd{}.
\item \emph{Image augmentations:} Both methods apply data augmentations to the image $x$ at each step in training.
However, the specific augmentations used in CLIP (\texttt{resize} and \texttt{crop}) differ from those used for \simclr (\texttt{resize}, \texttt{crop}, \texttt{flip}, \texttt{jitter}, \texttt{blur}, \texttt{grayscale}).
We remove this confounder by using standard \simclr augmentations unless otherwise specified.

\item \emph{Architecture:} We use the ResNet-50~\cite{he2016deep} architecture as the image encoder for both methods, and a Transformer~\cite{vaswani2017attention} as the text encoder (for captions) in CLIP.

\item \emph{Datasets:} Typically, CLIP and SimCLR are trained on different datasets, as the former requires matched image-caption pairs, while the latter can leverage any computer vision dataset.
To control for the effect of the data distribution, we pre-train both models on the same datasets: starting with the relatively controlled MS-COCO

\item \emph{Hyperparameters:} We extensively tune hyperparameters for both methods (Appendix~\ref{app:hyperparameters}). 
\end{itemize}

\paragraph{Mismatches.}
Despite our efforts to match CLIP and SimCLR, there are some inconsistencies that we are unable to account for---partly due to the differences in their modalities.
In particular, CLIP: 
\begin{itemize}
\item[(i)] Processes $T(x)$ using a text transformer rather than SimCLR's ResNet-50.
\item[(ii)] Does not share weights between the encoders processing $x$ and $T(x)$ because they correspond to different modalities, unlike SimCLR.
\item[(iii)]  Uses a linear projection head $g/g'$ instead of SimCLR's MLP, which we allow as \citet{radford2021learning} showed that this choice does not affect CLIP's performance.
\item[(iv)] Only uses other examples in the batch from the \emph{same} modality as negatives. Thus CLIP has half the number of negatives compared to SimCLR, which also uses transformed versions of other examples in the batch (\ie both $\hat{x}$ and $\hat{x}_+$) as negatives .
\end{itemize}

In Sections~\ref{sec:coco_study} and~\ref{sec:quality}, we take a closer look at how our matched versions of the CLIP and SimCLR methods compare in terms of downstream transfer performance.

\subsection{A COCO case study}
\label{sec:coco_study}
We begin our study by comparing CLIP and SimCLR models trained on the MS-COCO dataset~\citep{lin2014microsofta} (henceforth referred to as COCO), which contains $\sim$120K images with multi-object labels.
Each image has five human-provided captions, collected post-hoc by \citet{chen2015microsoft} using Amazon Mechanical Turk.
Annotators were given detailed instructions on how to caption an image such as to describe \emph{only} the important parts of the image, not to use proper names, and to use at least 8 words.

We use COCO as our starting point for two reasons.
First, we can assess the utility of language supervision in the ideal setting where the captions are of fairly high quality due to the careful curation process. 
Second, we can approximate \clipd{}\footnote{We overload notation and use \clipd{} to denote: (i) the idealized stochastic version of CLIP, which samples from infinite captions per image, and (ii) our approximation of it using a finite set of (typically five) image captions.} by sampling from the available set of five captions per image.

\input{tables/coco_clip_vs_simclr}

\paragraph{Captions (often) help on COCO.} 

In Table~\ref{tab:coco_comparison_basic}, we compare pre-trained models (SimCLR, CLIP, \clipd{} and a supervised baseline), in terms of the accuracy of a linear probe on: (i) COCO classification (in distribution), and (ii) transfer to downstream tasks from \citet{kornblith2019do}.

On COCO classification, supervised models outperform self-supervised ones, and \simclr is more accurate than  CLIP trained on human-written captions.
This trend, however, flips when we consider out-of-distribution performance. 
On most transfer tasks, CLIP performs the best: yielding on average 10\% accuracy over the supervised baseline and 2\% over SimCLR.

Using a stochastic version of CLIP further boosts its performance.
\clipd{} matches \simclr performance in-distribution and is about 5\% better on average in terms of transfer.
While this improvement is remarkable, it is not clear why stochastically sampling different (one-of-five) captions for a given image helps.
For instance, it may be optimization-related or linked to properties of the captions themselves.
We revisit this question in Section~\ref{sec:diversity}.

Notably, we find that matching the image augmentations (applied to $x$) is crucial to correctly assess the merit of added language supervision.
In particular, using standard CLIP augmentations (only \texttt{resize} and \texttt{crop}) during training lowers its average transfer accuracy by 10\% (Appendix Table~\ref{tab:app_coco_comparison_basic}).
With these same augmentations, SimCLR's performance also drops by 50\%.
This shows the importance of correctly controlling for potential confounders discussed in Section~\ref{sec:common_ground}.

%% file: tables/coco_clip_vs_simclr.tex

\begin{table}[!t]
	\centering
	\setlength{\tabcolsep}{2.5pt}
	\renewcommand{\arraystretch}{1.3}
	
\begin{tabular}{|l|c|cccccccccccc|c|}
\hline 
 & \zot{COCO} & \zot{Aircraft} & \zot{Birdsnap} 
 & 
 \zot{Ctech101} & \zot{Ctech256} & \zot{Cars} & 
 \zot{CIFAR10} & \zot{CIFAR100 \hspace{0.1ex}} & 
 \zot{DTD} & 
 \zot{Flowers} & \zot{Food-101} & \zot{Pets} & 
 \zot{SUN937} & 
$\mu_{\text{Tx}}$ \\ \hline
Supervised & \textbf{90.6} & 31.6 & 11.8 & 65.8 & 53.7 & 21.7 & 74.8 & 46.7 & 55.9 & 63.4 & 47.1 & 45.9 & 44.5 & 47.2 $\pm$ 0.2 \\
\rowcolor{LightRed}
\simclr & 89.0 & 40.6 & 18.5 & 71.5 & 58.6 & 31.5 & 82.1 & 57.3 & 61.7 & 77.4 & 58.7 & 57.3 & 51.9 & 56.0 $\pm$ 0.2 \\
\rowcolor{LightCyan}
CLIP & 88.4 & 41.4 & 17.6 & 73.2 & 60.4 & 35.8 & 83.6 & 60.8 & 65.7 & 80.5 & 60.9 & 57.0 & 50.8 & 57.5 $\pm$ 0.1 \\
\rowcolor{LightGreen}
\clipd & 89.8 & \textbf{46.4} & \textbf{20.0} & \textbf{78.4} & \textbf{65.6} & \textbf{41.5} & \textbf{84.6} & \textbf{62.5} & \textbf{66.7} & \textbf{83.9} & \textbf{65.3} & \textbf{61.2} & \textbf{54.9} & \textbf{61.3 $\pm$ 0.2} \\
\hline
\end{tabular}
\caption{Linear probe accuracy for COCO pre-trained models in-distribution and on the transfer suite from Kornblith et al.~\cite{kornblith2019do}. Here, $\mu_{Tx}$ denotes the average test accuracy of the model over downstream transfer tasks. We report 95\% confidence intervals (CI) via bootstrapping. }
	\label{tab:coco_comparison_basic}
\end{table}

%% file: sections/quality.tex
\section{The impact of pre-training data}
\label{sec:dp}
Our analysis of COCO shows that language supervision can indeed be beneficial over using images alone.
However, the datasets that CLIP is typically trained on differ, both in scale and quality, from COCO.
For instance, COCO captions were collected post-hoc under controlled settings, which is markedly different from the automatic scraping procedure used to gather data at scale.
Thus, we shift our focus to two more prototypical pre-training datasets:
\begin{itemize}
	\item \emph{ConceptualCaptions}~\citep{sharma2018conceptual} (CC) contains $\sim$3.3M images harvested from web, with their ALT-text HTML attributes as captions. The dataset was filtered (retaining only 0.2\%) for text quality---e.g., well-formed captions that mention at least one object found via the Google Cloud Vision API. Furthermore, all proper nouns in the captions were hypernymized (e.g., "Justin Timberlake" becomes "pop artist").
 	\item \emph{Yahoo Flickr Creative Commons}~\cite{thomee2016yfcc100m} (YFCC): This dataset has $\sim$ 99.2M images from Flickr, along with their posted titles as captions with no filtering or post-processing.
\end{itemize}

\begin{figure}[!t]
	\centering
	\includegraphics[width=0.975\textwidth]{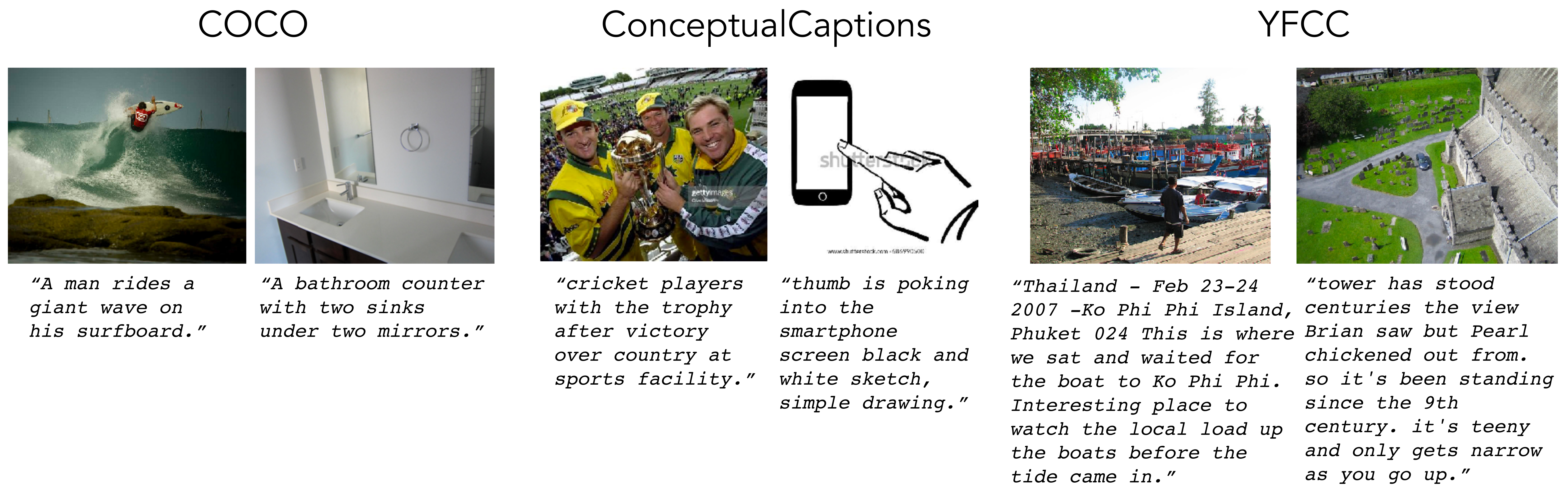}
	\caption{Random samples from the COCO, CC and YFCC datasets (also see Appendix Figure~\ref{fig:app_datasets}). There are noticeable differences in the diversity of their images and the corresponding captions.}
	\label{fig:datasets}
\end{figure}

We now assess whether our findings on COCO translate to the CC and YFCC datasets.
We start by comparing the transfer performance of CLIP and SimCLR on COCO-scale subsets of CC/YFCC---cf. points corresponding to 100K samples in Figure~\ref{fig:dataset_comparison} (\emph{right}).
We observe that SimCLR's performance does not vary much across pre-training datasets.
On the other hand, CLIP's transfer capabilities are highly sensitive to the dataset.
With 100K samples from CC/YFCC, using CLIP is \emph{worse} than pre-training only on images---in contrast to what we found on COCO.

Inspecting COCO, CC and YFCC samples (Figure~\ref{fig:datasets}) yields a possible explanation for this sensitivity.
The datasets differ not just in scale and image diversity, but also the extent to which their captions: (i) describe visually salient aspects of the image, and (ii) vary across images (\eg, in style and wording).
For instance, captions in COCO are homogenous and descriptive, while in YFCC, they vary and often complementary to the image.
In Sections~\ref{sec:scale}-\ref{sec:diversity}, we investigate the effect these properties (scale, descriptiveness and variability) of the dataset have on CLIP's performance.

\subsection{Scale matters}
\label{sec:scale}

A major advantage of contrastive learning methods is that they can leverage the vast amounts of unlabeled data available on the Internet.
Thus, it is natural to ask how different forms of contrastive supervision benefit from added pre-training data.
Intuitively, we expect image-only methods to perform worse for smaller datasets as they are increasingly less likely to encounter images with similar augmentations.
We might further expect image-language models to perform more favorably in this setting since they receive richer supervision.

\begin{figure}[!t]
	\centering
	\begin{subfigure}{0.475\textwidth}
		\centering
		\includegraphics[width=0.92\textwidth]{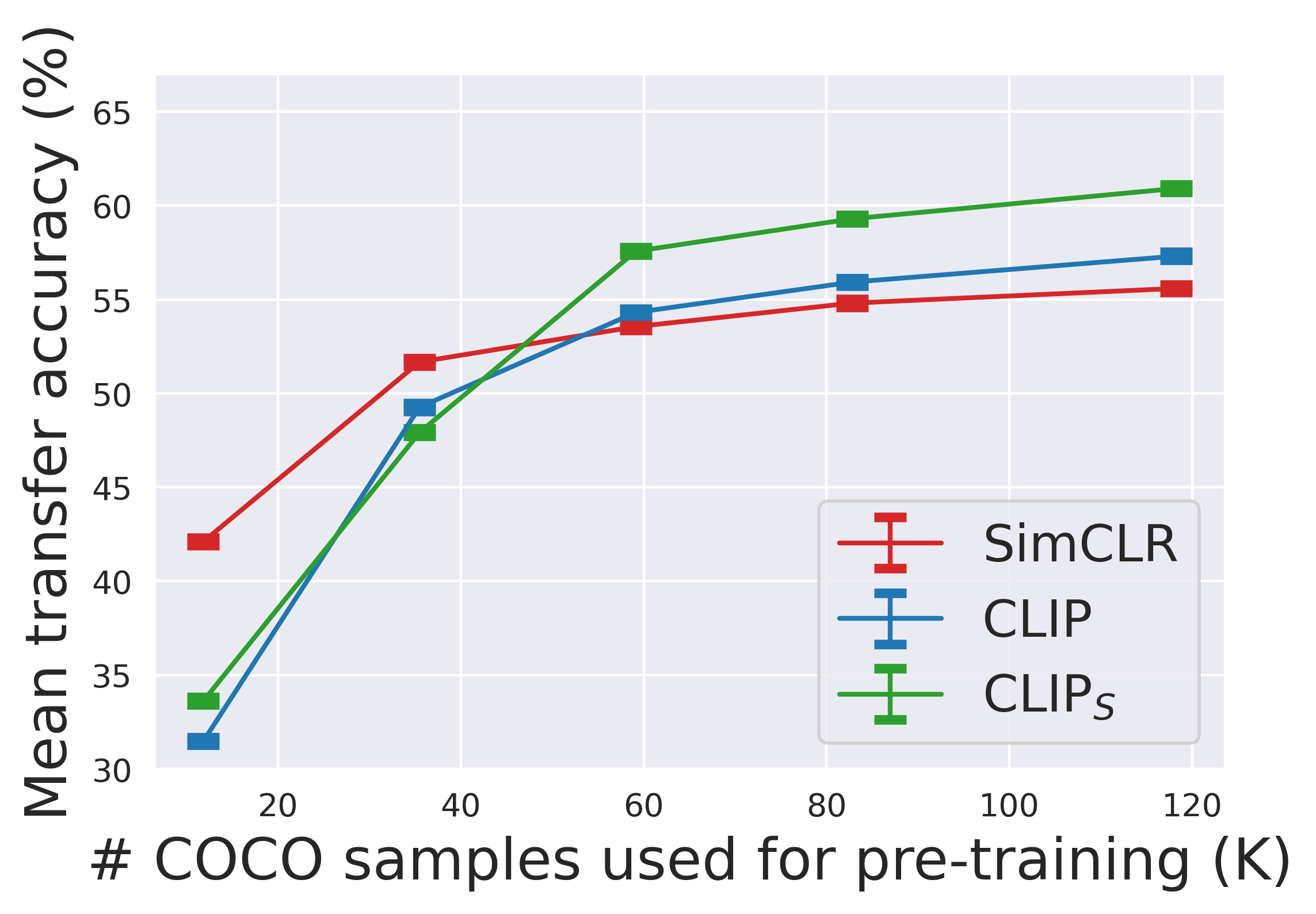}
	\end{subfigure}
	\hfil
	\begin{subfigure}{0.475\textwidth}
		\centering
		\includegraphics[width=0.92\textwidth]{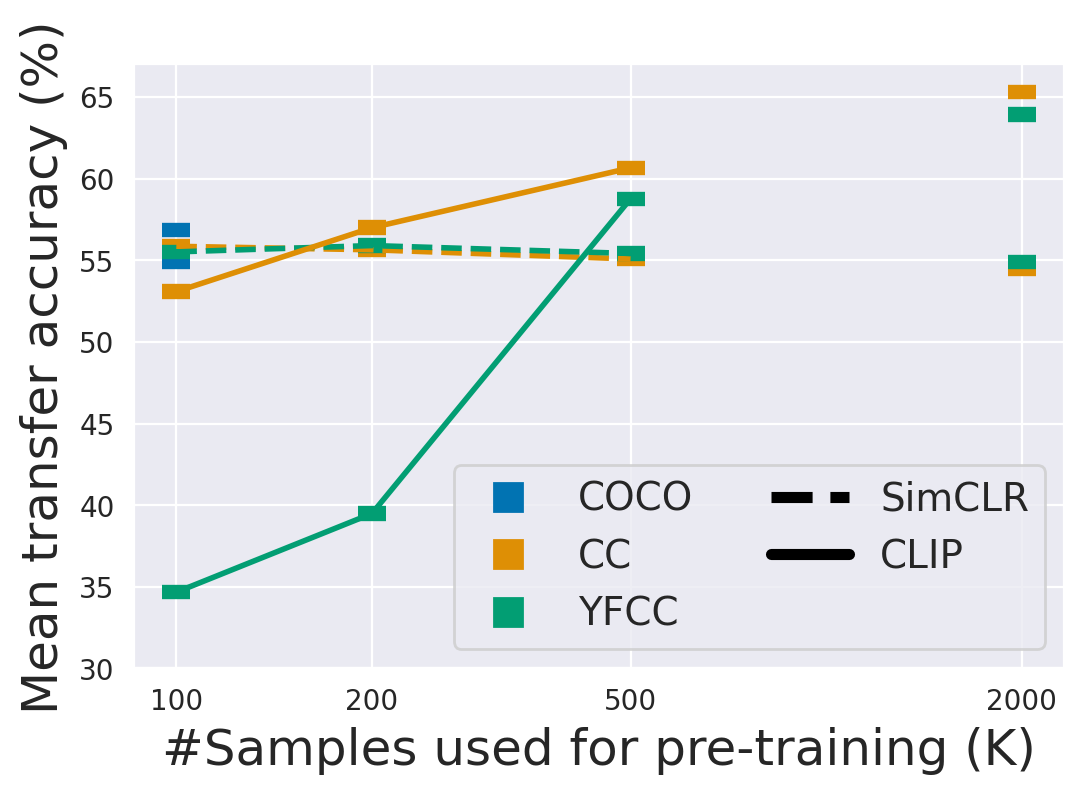}
	\end{subfigure}
	\caption{Average transfer accuracy of models w.r.t. pre-training dataset size for COCO (\emph{left}), and CC and YFCC (\emph{right}). 
	Language supervision consistently improves performance in the medium to large data regime over using images alone.
	However, on small corpora, providing the model with additional information via captions is actually detrimental. 
	Due to computational constraints, we train models for fewer epochs (100 instead of 200) on datasets of size 2M.}
	\label{fig:dataset_comparison}
\end{figure}

To test whether this is the case, we compare CLIP and SimCLR models trained on datasets of varying sizes: in the 10-100K sample regime for COCO, and 100K-2M sample regime for CC/YFCC.\footnote{Due to computational constraints, we train CLIP/SimCLR for fewer epochs (100 instead of 200) on 2M examples.} 
Our results in Figure~\ref{fig:dataset_comparison} deviate from our earlier expectations.
First, beyond a certain point, SimCLR's transfer performance improves only marginally with additional data.
While surprising, similar effects have been noted previously~\cite{tian2021divide,cole2022does}, especially when the data is uncurated (e.g., YFCC)~\cite{tian2021divide}.\footnote{Table~\ref{tab:app_justify} in the appendix shows that even more sophisticated image contrastive learning methods studied in \citet{tian2021divide}---trained with better data augmentations, with 4x batch size, on 50x the data and for 10x more epochs---are only marginally better than our CLIP models on the downstream tasks from \citet{kornblith2019do}.}
Second, in the low-data regime ($<$50K/200K/500K for COCO/CC/YFCC), training with language actually hurts the models' transfer performance.

In fact, we find that (data) scale is essential to take advantage of language supervision.
With sufficient data, CLIP outperforms SimCLR on all three datasets.
This gap remains even if we train SimCLR with extra data,
indicating that captions can be worth more than \emph{any} number of images.

\subsection{The importance of descriptive captions}
\label{sec:quality}

As we saw in Figure~\ref{fig:datasets}, captions in typical datasets can vary in terms of how they relate to the image.
Prior work in linguistics and accessibility has drawn a distinction between captions that are \emph{descriptive} (meant to replace an image) and \emph{complementary} (meant to give additional context)~\citep{hodosh2013framing,kreiss2021concadia,dognin2022image}. 
This line of work suggests that COCO captions are more {descriptive} due to the decontextualization of the image and strict instructions provided to the annotators during the caption generation process~\citep{kreiss2021concadia}.
In contrast, Flickr captions (\eg, in YFCC) tend to contain information that is complementary to the image since people typically do not restate what can already be observed in the photographs they post~\cite{grice1975logic}.

In representation learning for object recognition tasks, we ideally want to meaningfully encode salient objects in the image.
Recall that for contrastive models the learned representations are determined by the transformation $T(x)$ (captions for CLIP).
This suggests a hypothesis: captions that describe the contents of a scene will improve CLIP's transferability.

To test this hypothesis, we need to quantify the {descriptiveness} of a caption.
Since measuring it precisely is infeasible, we approximate it with the help of a pre-trained caption-scoring model (BLIP~\cite{li2022blip}).
Specifically, we use the score given by BLIP of a caption matching its corresponding image as a surrogate for descriptiveness.
Comparing the average caption descriptiveness of the three datasets in Figure~\ref{fig:quality_comparison_all}, we see that COCO $>$ CC $>$ YFCC. 
This aligns with our earlier subjective assessment as well as with prior work~\cite{hodosh2013framing,kreiss2021concadia}.	

In Figure~\ref{fig:quality_comparison_all}, we visualize the relationship between the average descriptiveness of a dataset's captions and the transfer performance of the resulting CLIP model. 
We indeed find that descriptive captions are crucial for CLIP's performance, and one descriptive image-caption pair from COCO is worth 2x and 5x samples from CC and YFCC respectively. 
On YFCC and CC, CLIP thus requires more data to benefit from language supervision.

Finally, we train CLIP on 100K subsets of CC and YFCC with ``more descriptive'' captions by re-captioning the images using BLIP~\citep{li2022blip} (examples in Appendix Figure~\ref{fig:app_blip_examples}).
CLIP trained on CC/YFCC using BLIP captions no longer performs worse than its COCO counterpart (Figure~\ref{fig:quality_comparison_all}). 
This indicates that CLIP's sensitivity to the pre-training corpus is not just an artifact of differing image distributions, but due to the presence (or absence) of descriptive captions.

\input{tables/coco_div}

%% file: tables/coco_div.tex
\begin{figure}[!t]
	\centering
	\setlength{\tabcolsep}{2.5pt}
	\renewcommand{\arraystretch}{1.4}
		\begin{subfigure}{0.475\textwidth}
		\centering
        \includegraphics[width=0.85\textwidth]{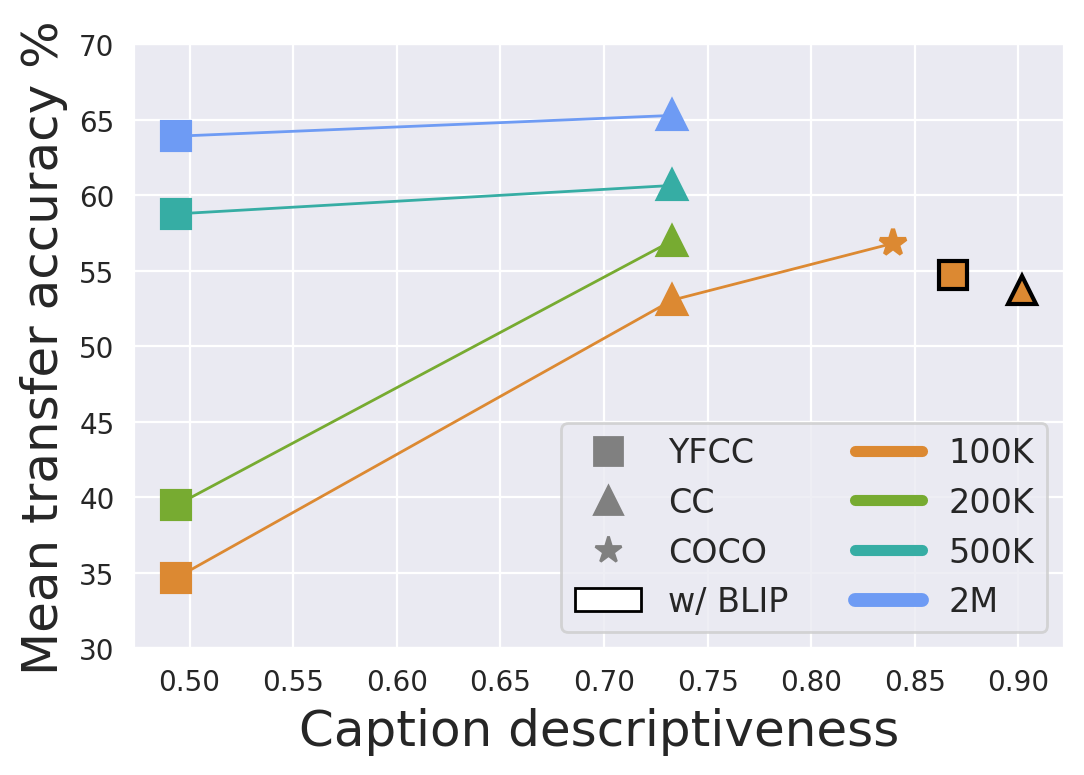}
		\caption{}
		\label{fig:quality_comparison_all}
	\end{subfigure} \hfill
	\begin{subfigure}{0.475\textwidth}
		\centering
	
		\begin{tabular}{|ccc|cc|}
		\hline Cons. & Comp. & Model  
		& {COCO} & {$\mu_{Tx}$} \\ \hline
		\checkmark & \checkmark &  CLIP & 88.8 & 59.2 $\pm$ 0.1 \\ \hline
		\checkmark & \xmark & CLIP &  88.4 &  57.7 $\pm$ 0.2 \\
		\rowcolor{Gray}
		\checkmark & \xmark & \clipd{} & 89.1  & {59.3 $\pm$ 0.2} \\\hline
		\xmark & \xmark & CLIP &  88.4  & 56.6 $\pm$ 0.2 \\
		\rowcolor{Gray}
		\xmark & \xmark & \clipd{} & {89.3} & 58.9 $\pm$ 0.2 \\
		\hline 
		\end{tabular} \vspace{1.5em}
		\caption{}
		\label{fig:coco_diversity}
	\end{subfigure}
	\caption{Relationship between CLIP's transfer performance and ({a}) the average \emph{descriptiveness} of dataset captions and ({b}) intra-dataset caption \emph{\var}.
	We approximate descriptiveness (a) using a pre-trained BLIP model~\cite{li2022blip} to score the similarity between a caption and the image it corresponds to.
	To study the effect of caption variability in (b), we construct synthetic captions for the COCO dataset using its multi-object image labels.
	We vary whether these captions are consistent (use a single term to describe a given object) and complete (describes all image objects). }
	\label{fig:caption_properties}
\end{figure}

%% file: sections/diversity.tex
\subsection{The effect of intra-dataset variations in captions}
\label{sec:diversity}

Next, we examine how the variability of captions within a dataset affects CLIP's transfer capabilities.
After all, there are many ways to caption an image, as shown in Figure~\ref{fig:intro_fig}.
The presented captions vary in terms of how they describe an object (e.g., ``duffel van'' or ``car''), and the parts of the image they focus on (e.g., discussing the ``street'' or ``brick'').
These stylistic, lexical, and focus variations in captions could make it harder for CLIP to learn meaningful representations.

\paragraph{A simple setting.} We investigate this effect on the COCO dataset by creating synthetic captions using multi-object image labels (examples in Appendix Figure~\ref{fig:app_synthetic_examples}).
Here, we can construct captions to precisely control whether they are: 
(i) \emph{consistent:} by either using a fixed term or random synonyms to describe an object across the dataset; and 
(ii) \emph{complete:} by either mentioning all or a random subset of image objects.
In this setting, we find:

\begin{itemize}
    \item A CLIP model trained with complete and consistent synthetic captions \emph{outperforms} a model trained on human-written captions (cf. row 1 in Figure~\ref{fig:coco_diversity} to row 3 in Table~\ref{tab:coco_comparison_basic}).
    \item Dropping these two conditions, and thereby increasing the variability of captions, causes the transfer performance of the model to drop (cf. rows 1, 2, and 4 in Figure~\ref{fig:coco_diversity}).
    \item A stochastic version of CLIP, \ie \clipd{} based on 5 synthetic captions per image, is not as affected by caption inconsistency and/or incompleteness. The $\sim$2\% improvement of \clipd{} over CLIP here mirrors the 3.6\% gain seen for human-provided captions in Figure~\ref{tab:coco_comparison_basic}. 
    \item Unlike CLIP, \clipd{} transfers 2\% better on average when trained on human-provided captions as opposed to synthetic ones.
\end{itemize}

\noindent 
These findings suggest that variability in dataset captions does have an adverse effect on the resulting CLIP models.
This drop can be mitigated by stochastically sampling from a set of possible captions per image, rather than using a single fixed caption (in \clipd).
Note that standard image contrastive learning methods already do this since they use random data augmentations (\eg, a different crop) to generate $T(x)$ at every epoch.
Finally, our results show that human-written captions contain useful information for representation learning that is not present in object labels alone.
However, extracting this signal is not straightforward, and may require incorporating multiple captions into CLIP training.

\paragraph{Datasets in practice.}
Looking at Figure~\ref{fig:datasets}, it seems that COCO $<$ CC $<$ YFCC in terms of caption variability.
This trend may be expected given how their captions were obtained and/or post-processed: careful post-hoc labeling for COCO, filtering and hypernymizing ALT-text for CC, and scraping raw Flickr titles for YFCC.
Our results in the simple setting above suggests that this variability of YFCC captions (and to a lesser extent CC) could (with lower descriptiveness) be responsible for the worse transfer performance of the resulting CLIP models (Figure~\ref{fig:dataset_comparison} right).
It also explains why scale is essential to benefit from language supervision on CC and YFCC.
After all, CLIP would need to be trained on more captions to even encounter the same word twice.



\input{tables/blip}

\paragraph{How many captions are enough?}
In the simple setting above, we saw that performing ``text data augmentations'' in \clipd{} can reduce the adverse impacts of caption variability.
We now analyze how this effect scales with the number of available captions per image, focusing on the CC and YFCC datasets. 
Since these datasets only contain one caption per image, we use the BLIP captioning model to generate multiple captions via nucleus sampling~\cite{holtzman2020curious}.
We observe in Figure~\ref{fig:blip_diversity_left}, that \clipd{} improves as the number of available (BLIP-generated) captions per image increases (plateauing around 10).
However, scaling up the overall number of image-caption pairs appears to be far more effective than incorporating more captions per image (at least those obtained via BLIP) from the perspective of improving transfer performance (see Figure~\ref{fig:blip_diversity_right}).
Note that the relative costs of these two approaches are context dependent---\eg in Section~\ref{sec:prescription} we discuss ways to augment the pool of image captions in a dataset without additional data collection. \\


\noindent It is important to note that there is a strong inter-dependence between the three properties we discussed in Section~\ref{sec:dp} in terms of their influence on CLIP's transfer capabilities.
For instance, scale can, to an extent compensate for variable/less descriptive captions (as seen in Figure~\ref{fig:dataset_comparison}).
That being said, our findings suggest that the utility of captions as a form of supervision can be greatly improved by being more mindful of \emph{what} and \emph{how} they describe (in) an image. 

%% file: tables/blip.tex
\begin{figure}

	\centering
    \setlength{\tabcolsep}{2.5pt}
	\renewcommand{\arraystretch}{1.3}
    \begin{subfigure}{0.475\textwidth} \vspace{1em}
        \centering
        \begin{tabular}{|lcc|cc|} \hline
            Method & $N_C$ & Source  & CC (100K) & YFCC (100K) \\ \hline
            \simclr & 0 & -  & 55.9 $\pm$ 0.2  & 55.5 $\pm$ 0.2 \\
            CLIP & 1 & Human & 53.1 $\pm$ 0.2 & 34.7 $\pm$ 0.2 \\
            CLIP & 1 & BLIP & 53.7 $\pm$ 0.2 & 54.8 $\pm$ 0.2 \\
            \clipd & 2 & BLIP & - & 56.9 $\pm$ 0.2 \\
            \clipd & 5 & BLIP & {57.8 $\pm$ 0.2} & 58.8 $\pm$ 0.2\\
            \clipd & 10 & BLIP &  - & {59.1 $\pm$ 0.2} \\
            \hline
        \end{tabular}   \vspace{1em}
        \caption{}
        \label{fig:blip_diversity_left}
    \end{subfigure}
    \hfill
    \begin{subfigure}{0.475\textwidth}
        \centering
        \includegraphics[width=0.82\textwidth]{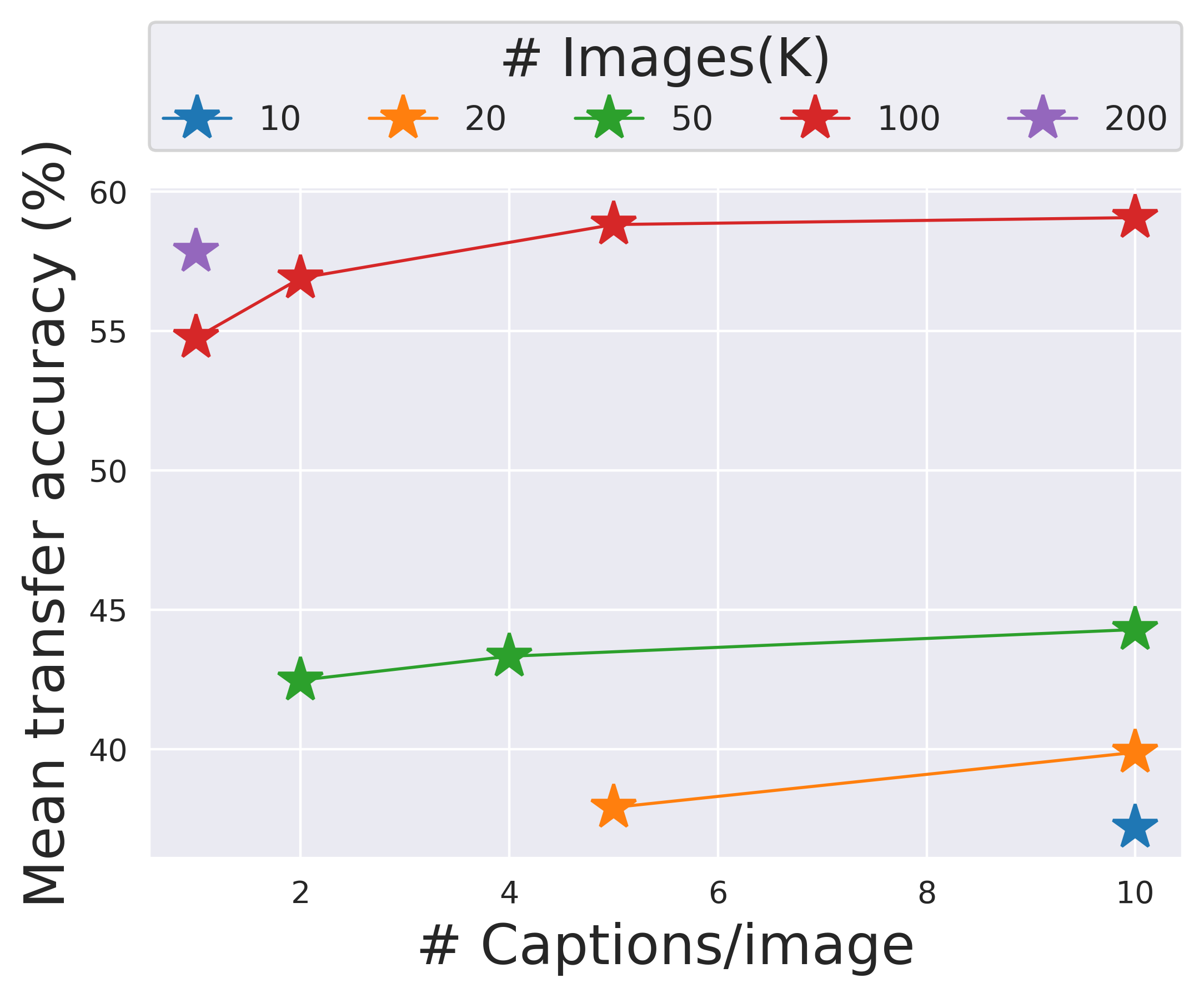}
        \caption{}
        \label{fig:blip_diversity_right}
    \end{subfigure}
    \caption{A closer look at \clipd{}. ({a}) Sensitivity of its transfer performance to the number of captions per image ($N_C$) used during training. For CC and YFCC, we use the BLIP captioning model to generate multiple diverse captions per image. ({b}) Performance trade-offs between pre-training using more image-caption pairs vs. more captions per image on the YFCC dataset.}
    \label{tab:blip_diversity}
\end{figure}

%% file: sections/prescription.tex
\section{Making existing captions work}
\label{sec:prescription}

So far, we have identified two properties of captions that impact their usefulness as a mode of supervision: (i) descriptiveness and (ii) variability.
With these in mind, we now put forth simple interventions that can be made to datasets to improve the performance of CLIP models.

\paragraph{Data pre-processing:}
Given the importance of caption descriptiveness, we might consider pre-processing scraped data to select for samples with this property.
The CC data collection procedure~\citep{sharma2018conceptual} partially demonstrates the effectiveness of this approach, as pre-training CLIP on CC samples leads to better transfer performance than a comparable number of ``raw'' YFCC ones. 
However, due to its reliance on the Google Vision API, this procedure can be quite expensive, with costs scaling with the size of the scraped data.
Recent works have taken a different approach, using pre-trained image-language matching models (like CLIP and BLIP) to filter data~\cite{schuhmann2021laion-400m*a}.
Given that we are interested in building such models in the first place, we avoid taking this route.

Instead, we focus on understanding how far we can get by simply discarding low quality captions, agnostic to the images.
To do so, we take inspiration from the filtering pipelines used to build large language models~\cite{brown2020language}.
Here, raw {Internet} data is cleaned by selecting samples that are ``similar'' to known high-quality datasets (\eg, Wikipedia). 
Taking a similar approach, we train a linear classifier on a bag-of-n-grams sentence embeddings~\cite{joulin2017bag} to distinguish validation set CC/YFCC captions from COCO ones.
This classifier is then used to filter CC/YFCC, only retaining  samples that are predicted as being COCO-like (examples in Appendix Figure~\ref{fig:app_fasttext_examples}).
For a given pre-training data budget, we see moderate gains ($\sim2\%$) from using this simple heuristic to filter the CC and YFCC datasets---see Table~\ref{tab:intervention} (\emph{left}).
To put these gains in context, we also report the performance of CLIP trained on the same images with ``high-quality'' BLIP-generated captions.

\paragraph{Mitigating caption variability:} 
As we saw in Section~\ref{sec:diversity}, models trained with \clipd{} are less impacted by caption variability.
However, typical image-caption datasets (such as CC and YFCC) only have one caption per image.
We thus devise a methodology to augment these captions by leveraging recent open-source large language models~\cite{wang2021gpt-j-6b}.
Concretely, we provide GPT-J with $4$ (caption, paraphrase) pairs as \emph{in-context}~\cite{brown2020language} examples.
We then prompt it to paraphrase a given target caption.
By sampling from GPT-J, we can obtain multiple (in our case five) paraphrases for every such caption (see Appendix Figure~\ref{fig:app_gptj_examples} for examples).
In Table~\ref{tab:intervention} (\emph{right}), we see that feeding these captions into \clipd{} results in a considerable performance boost over CLIP (trained with a single caption/image).
For instance, for COCO, \clipd{} trained on our generated captions bridges more than half of the performance gap between CLIP and \clipd{} trained with one and five human-provided captions respectively.

\input{tables/filter}



%% file: tables/filter.tex
\begin{table}[!t]
	\begin{subtable}[h]{0.45\textwidth}
		\label{tab:filter}
		\centering
		\setlength{\tabcolsep}{2pt}
		\renewcommand{\arraystretch}{1.4}
		\begin{tabular}{|clc|c|}
		\hline  Dataset & Method & Preproc. & {$\mu_{Tx}$} \\ \hline
		\multirow{3}{*}{CC (100K)} & \simclr{} & - & 55.9 $\pm$ 0.3 \\
		&  CLIP & - & 53.1 $\pm$ 0.2 \\ 
		\rowcolor{Gray}
		&  CLIP & Filter & 54.2 $\pm$  0.2 \\  \hline
		\multirow{3}{*}{YFCC (500K)} & \simclr{} & - & 55.4 $\pm$ 0.2  \\
		&  CLIP & - & 58.8  $\pm$   0.2 \\ 
		&   CLIP & BLIP & {61.8 $\pm$  0.2} \\ 
		\rowcolor{Gray}
		&   CLIP & Filter & 60.4 $\pm$ 0.2 \\
		
		\hline
		\end{tabular}
	\end{subtable} \hfil
	\begin{subtable}[h]{0.45\textwidth}
		\label{tab:gpt}
		\centering
		\setlength{\tabcolsep}{3pt}
		\renewcommand{\arraystretch}{1.4}
		\begin{tabular}{|clc|c|}
		\hline  Dataset  & Method & Caption & {$\mu_{Tx}$} \\ \hline
		\multirow{4}{*}{COCO (120K)} & \simclr{} & - & 56.0 $\pm$ 0.2 \\
		&  CLIP & Human & 57.5 $\pm$ 0.1  \\
		&   \clipd{} & Human & 61.3 $\pm$ 0.2  \\ \rowcolor{Gray}
		&   \clipd{} & GPT-J & 58.9 $\pm$ 0.3\\ \hline
		\multirow{3}{*}{CC (200K)} &  \simclr{} & - & 55.3 $\pm$ 0.2 \\
		&  CLIP & Human & 57.0 $\pm$ 0.3 \\ \rowcolor{Gray}
		&   \clipd{} & GPT-J & {58.8 $\pm$ 0.3} \\
		\hline
		\end{tabular}
		\end{subtable}
	\caption{Improving CLIP's transfer performance through simple interventions on existing datasets. (\emph{left}) Applying a simple bag-of-words classifier to identify data subsets with ``high quality'' captions. (\emph{right}) Using in-context learning with GPT-J to obtain five diverse captions for dataset images (via paraphrasing) which are then used to train \clipd{}. For COCO, we also compare to \clipd{} trained with five human-written caption.}
	\label{tab:intervention}
\end{table}

%% file: sections/related.tex
\section{Related Work}
\label{sec:related}

\paragraph{Representation learning.}
Building models with \emph{general} representations that transfer to downstream tasks has been a long-standing goal in ML~\cite{donahue2014decaf,razavian2014cnn,chatfield2014return,agrawal2014analyzing,yosinski2014how}.
Our work is in line with prior studies aimed at characterizing the effect of design choices made during training~\cite{azizpour2015factors,huh2016what,chu2016best,kornblith2019do,zhai2019large-scale,locatello2020sober}, \eg model architecture, datasets and loss functions, on learned representations.

\paragraph{The utility of language in vision.}
There has been a long line of work on leveraging language to improve vision models~\citep{quattoni2007learning,srivastava2012multimodal,frome2013devise,baltrusait2018multimodal,guo2019deep}.
However, with the development of CLIP and its variants \citep{mu2021slip,li2022supervision,yao2022filip}, this approach has become a serious contender to traditional image-only ones.
Follow up works have sought to investigate how integral language is to CLIP's performance.
\citet{ruan2022optimal} suggest theoretically that the robustness of linear probes on CLIP's representations stems from pretraining with a large and diverse set of images and domain-agnostic augmentations $T(x)$ . 
More recently, \citet{fang2022data} examined CLIP's effective robustness~\citep{taori2020measuring} (on ImageNet-like datasets~\cite{deng2009imagenet,russakovsky2015imagenet,recht2019do,wang2019learning,barbu2019objectnet,hendrycks2021natural,hendrycks2021many}) in the zero-shot setting.
They find that CLIP's robustness is comparable to that of a supervised classifier trained on the same pool of YFCC images, and therefore conclude that data distribution is more important than language supervision.
Our work is complementary to this study, as we examine the role of language in a different setting, \ie, self-supervised representation learning.
We show that the impact of language supervision in this setting is complex and depends on the quality and quantity of image-caption data.

Most similar to our work is the recent study by \citet{devillers2021does}, which argues that language supervision does not result in improved downstream transfer, few-shot learning and adversarial robustness.
While they also consider CLIP and image-only models (e.g. BiT \citep{kolesnikov2020big}), they do not attempt to directly control confounding effects.
In particular, the two sets of models are trained on different datasets with different objectives (\eg, contrastive for CLIP, supervised for BiT).
Our work performs a substantially more controlled study on the effect of language supervision, allowing us to make more direct claims than \citet{devillers2021does}.

\paragraph{Supervision in self-supervised learning.}
Prior works in contrastive learning have studied how properties of the transformation $T(x)$ affect the transferability of learned representations.
They show that for a given image $x$, a good view ($x_+$) is one that retains label information while removing other nuisances \citep{tian2020what,tsai2021self-supervised,dubois2021lossy,federici2020learning,mitrovic2021representation,wu2022on}.
From this perspective, our work can be viewed as studying whether a caption provides a better view for a given image compared to standard image augmentations.

%% file: sections/conclusion.tex
\section{Discussion}
\label{sec:conclusion}

Our work takes a step towards resolving the debate as to whether multi-modality, and language in particular, can improve visual representation learning. 
A comparison of CLIP with matched image-only \simclr models reveals that neither form of supervision (using images alone or coupled with language) is strictly better than the other.
Indeed, there are practical regimes where CLIP's performance cannot be matched using \simclr with \emph{any} amount of image data and others where language supervision is harmful.
This is a direct consequence of CLIP's sensitivity to its pre-training data, especially its scale, descriptiveness, and \var{} of the captions.
Through our analysis, we also discovered algorithmic improvements (\clipd)  and dataset modifications (filtering and augmenting captions) to better take advantage of language supervision.

%% file: sections/impact.tex
\paragraph{Limitations.}
Our exploration allows us to quantify the utility of language supervision (over using images alone) in a specific setting: transfer learning via probing on certain object recognition tasks~\cite{kornblith2019do}.
We view expanding the scope of our analysis as a direction for future work.
Further, despite the significant steps we took to control the differences between CLIP and SimCLR, there are still some  inconsistencies that have not been accounted for (discussed in Section~\ref{sec:apples}). 
Nevertheless, the differences between our and previous results \citep[e.g,][]{devillers2021does} suggest that we successfully pinned down some crucial confounders (architecture, augmentations, stochasticity, datasets, hyperparameters).
Finally, while we show that CLIP's representations are influenced by what the captions they are trained on describe, we sidestep whether or not this is always desirable.
After all, recent studies~\cite{birhane2021multimodal} show that vision-linguistic datasets have various biases and stereotypes, which we might not want our models to learn.


%% file: sections/acknowledgements.tex
\section*{Acknowledgements}

We are grateful to Niladri Chatterji, Elisa Kreiss, Nimit Sohoni and Dimitris Tsipras for helpful discussions.
SS is supported by Open Philanthropy, YD by a Knights-Hennessy Scholarship, and RT by the NSF GRFP under Grant No. DGE 1656518. We also thank Stanford HAI for a Google Cloud credits grant.

%% file: appendix/setup.tex
\section{Setup and Experimental details}
\label{app:setup}

\subsection{Datasets}
\label{app:datasets}

In Appendix Figure~\ref{fig:app_datasets}, we present (random) samples from the MS-COCO~\cite{lin2014microsofta}, Conceptual Captions~\cite{sharma2018conceptual} and YFCC datasets~\cite{thomee2016yfcc100m}.
We use the 2017 version of COCO, which contains five human-written captions along with multi-object image labels for each image.

\input{figures/std_examples}

\paragraph{Licenses.}
These datasets were obtained by scraping images from online hosting services (\eg Flickr).
Thus, the ownership of the images lies with the respective individuals that uploaded them. 
Nevertheless, as per their terms of agreement, the images can be used for research purposes.

\clearpage

Like most large-scale datasets, COCO, CC and YFCC have not been extensively vetted, and may contain identifying information or offensive content.
Characterizing the pervasiveness of these issues is an important and active area of research.
That being said, we do not redistribute the data, our work is unlikely to significantly further the risks from these datasets. 

\subsection{Models}
\label{app:models}

We rely on existing open source implementations of CLIP~\cite{ilharco_gabriel_2021_5143773} and SimCLR~\cite{falcon_2019_lightning} for all our experiments, with a ResNet-50 image encoder (feature dimension=2048), and a linear and MLP projection head respectively.
We use the transformer architecture from \citet{radford2021learning} for encoding captions in CLIP.
Unless otherwise specified, we use five captions per image to train \clipd.
For downstream transfer, we train a linear probe using task data.

\subsection{Hyperparameters}
\label{app:hyperparameters}
We ran an extensive hyperparameter grid for CLIP and \simclr  on MS-COCO and used the same configuration in the rest of our experiments. 
These defaults are stated in Appendix Table~\ref{tab:app_hyperparams}.

\begin{table}[!h]
	\centering
	\setlength{\tabcolsep}{4pt}
	\renewcommand{\arraystretch}{1.4}
	
\begin{tabular}{|c|ccccc|}
\hline
Model &  Batch Size & Epochs & Warmup & lr & wd \\ \hline
Supervised & 1024 & 200 & 10 & $10^{-3}$ & $10^{-6}$  \\
\simclr & 1024 & 200 & 10 & $10^{-2}$ & $10^{-6}$  \\
CLIP & 1024 & 200 & 10 &  $10^{-3}$ & 0.1  \\ \hline
\end{tabular}
\caption{Default hyperparameters for model training.}
\label{tab:app_hyperparams}
\end{table}

We use the Adam optimizer with a cosine lr schedule for all the models.
All other hyperparameters are defaults from standard implementations of \simclr\footnote{\url{https://pytorch-lightning-bolts.readthedocs.io/en/latest/self_supervised_models.html}} and CLIP.\footnote{\url{https://github.com/mlfoundations/open_clip}}

\paragraph{Exceptions.}
We train CLIP/SimCLR on CC/YFCC-2M for 100 epochs due to computational restrictions.
For corpora smaller than 100K (Figure~\ref{fig:dataset_comparison}), we scale up the number of epochs to keep the number of iterations roughly comparable.

\paragraph{Data augmentations.}
The PyTorch pseudo-code for the default \simclr and CLIP data from prior work augmentation are as follows:
\begin{eqnarray*}
    T_{SimCLR} & = & \{ \texttt{RandomResizedCrop}(\texttt{size}=224), \\ \nonumber
                & &  \  \texttt{RandomHorizontalFlip}(\texttt{p}=0.5) \\ \nonumber
                & & \  \texttt{RandomApply}(\texttt{ColorJitter}(
                    0.8,
                    0.8,
                    0.8,
                    0.2
                ), \texttt{p}=0.8) \\ \nonumber
                & & \ \texttt{RandomGrayscale}(\texttt{p}=0.2), \\ \nonumber
                & & \ \texttt{GaussianBlur}(\texttt{kernel\_size}=23, \texttt{p}=0.5) \} \nonumber
\end{eqnarray*}
\begin{eqnarray*}
    T_{CLIP} & = & \{ \texttt{RandomResizedCrop}(\texttt{size}=224, \\ \nonumber
   & & \ \texttt{scale}=(0.9, 1.0), \\ \nonumber
   & & \ \texttt{interpolation}=\texttt{BICUBIC}) \} \nonumber
\end{eqnarray*}
Note that for our experiments, unless otherwise specified, we use the standard SimCLR set for the Supervised/SimCLR/CLIP models.

\paragraph{Linear probe.}
We train the probe using cross-entropy loss on CLIP/SimCLR features of dimensionality 2048.
In cases where the downstream task data is imbalanced, we re-weight the loss to account for it.
We also evaluate class balanced accuracy at test time.
For each downstream task, we train the probe for 250 epochs using an SGD optimizer.
We use a batch size of 256, weight decay of $10^{-6}$ and momentum $0.9$.
We perform a grid search for the best learning rate (using the validation set), considering values between 3$\times 10^{-2}$ and 10.
We also consider 3 random seeds.

\paragraph{COCO supervised.}
The COCO dataset contains multi-object labels for each image. 
We thus train the supervised classifier and linear probe in this setting to predict whether each of the 80 objects is present in an image.
We then evaluate the accuracy of the model by aggregating (in a class balanced manner) the correctness of each of these binary predictions.

\paragraph{Confidence intervals.}
We report 95\% confidence intervals obtained via bootstrapping over the test set, as well as the three random seeds used for the linear probe.
Due to space constraints, we do not always report them in the main paper, but include a detailed table for all our experiments with confidence intervals in the Appendix.

\subsection{Compute}
\label{app:compute}
We train each of our models of 4 NVIDIA A100 GPUs.
Training both CLIP and \simclr models takes on the order of 8-10 hours for a pre-training corpus of size $\sim$100K.

\newpage

\subsection{BLIP recaptioning}
\label{app:blip_recaption}
To generate BLIP captions for images from the CC and YFCC datasets, we use the BLIP captioning model~\cite{li2022blip}. 
In particular, we use the provided\footnote{\url{https://github.com/salesforce/BLIP}} ViT-Base with nucleus sampling  ($\text{top}_p$ of 0.9, repetition penalty of 1.1, and text length range [5, 40]), varying the random seed to generate multiple captions per image.
(Random) image-BLIP caption pairs are shown in Appendix Figure~\ref{fig:app_blip_examples}.
\input{figures/blip_examples}

\newpage

\subsection{Synthetic COCO captions}
\label{app:coco_synthetic}
We present examples of synthetic captions for the MS-COCO dataset created using the available multi-object image labels in Appendix Figure~\ref{fig:app_synthetic_examples}.
A synthetic caption is complete (incomplete) if it describes all (a random subset) of objects in the image.
It is consistent (inconsistent) if it describes a given object using a single consistent term throughout the dataset (one from a set of manually curated synonyms) and whether we use a fixed template (one of a set of templates).
In every case, we randomly order the objects that we describe.

\input{figures/synth_examples}

\subsection{Filtering captions}
\label{app:fasttex}

In Section~\ref{sec:prescription}, we introduce a methodology to filter poor quality captions from a given source dataset. 
Using the fastText library,\footnote{\url{https://github.com/facebookresearch/fastText}} we train a linear classifier bag-of-n-grams sentence embeddings (n=2) to distinguish a subset of source captions from those in the COCO validation set.
We then use the classifier to filter the source dataset, only selecting the ones that are (mis)classified as being COCO like.
In Appendix Figure~\ref{fig:app_fasttext_examples}, we present a (random) subset of filtered examples from the YFCC dataset.
Compared to random YFCC samples (cf. Appendix Figure~\ref{fig:app_datasets}), the ones in Appendix Figure~\ref{fig:app_fasttext_examples} have much shorter captions---often without attributes such as dates, urls and hashtags.
That being said, it is important to recognize that any simple heuristic for  filtering is ultimately limited by the captions present in the source dataset.

\input{figures/filt_examples}

\subsection{Augmenting captions with GPT-J}
\label{app:gptj}
We also propose a methodology to augment captions contained in existing datasets by using a pre-trained large language model (in our case GPT-J-6b, referred to as GPT-J) to paraphrase them (Section~\ref{sec:prescription}).
To this end, we rely on in-context learning, wherein we provide GPT-J with some (four) paired caption-paraphrase examples (using the five human-provided COCO captions) as the context.
We then ask GPT-J to paraphrase a given target caption.
For instance, a query to the model might look like:
\newpage

\begin{itemize}
    \item[] \texttt{Paraphrase the sentence below} 
    \item[] \texttt{Input: A little boy standing next to a dog in a field.}
    \item[] \texttt{Output: A dog parked filled with people and a bunch of different dogs.}
    \item[] \texttt{}
    \item[] \texttt{Paraphrase the sentence below}
    \item[] \texttt{Input: Some people are on the sandy beach flying kites.}
    \item[] \texttt{Output: a sunny day at the beach with colorful kites in the sky}
    \item[]
    \item[] \texttt{Paraphrase the sentence below}
    \item[] \texttt{Input: A living room filled with furniture and a table.}
    \item[] \texttt{Output: A living room with a nice couch and a coffee table.}
    \item[]
    \item[] \texttt{Paraphrase the sentence below}
    \item[] \texttt{Input: A couple of people on a surfboard in the ocean.}
    \item[] \texttt{Output: A dog is lying on the surfboard as it surfs on a wave.}
    \item[]
    \item[] \texttt{Paraphrase the sentence below}
    \item[] \texttt{Input: a handful of snowmen ~ make this with a glove .}
    \item[] \texttt{Output:}
\end{itemize}

We use temperature sampling to generate multiple diverse captions for a given image-caption pair from the dataset.
Examples for the CC dataset are shown in Appendix Figure~\ref{fig:app_gptj_examples}. 
\input{figures/gpt_examples}


\clearpage

%% file: figures/std_examples.tex
\begin{figure}[!h]
    \begin{minipage}[c]{0.3\hsize}\centering
        \centering
        \includegraphics[width=1.75in]{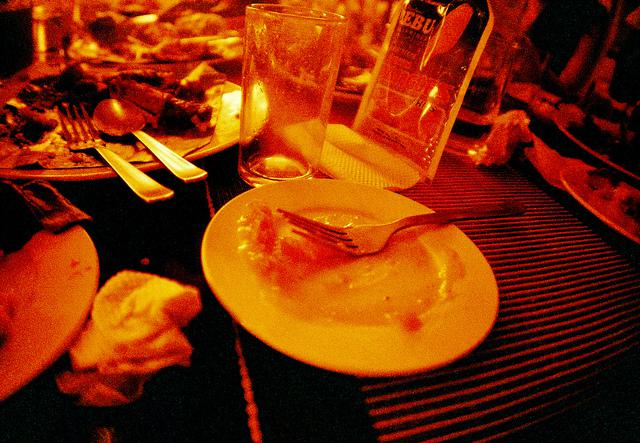}
        \caption*{}
    \end{minipage}
    \begin{minipage}[c]{0.75\hsize}\centering
        \begin{itemize}
            \setlength\itemsep{0.1em}
            \vspace{-2em}
            \item[-] ``A table topped with plates and glasses with eating utensils..''
            \item[-] ``a fork is laying on a small white plate''
            \item[-] ``dirty dishes on a table, and a bottle of something.''
            \item[-] ``a table top with some dishes on top of it'', 
            \item[-] ``A table full of dirty dishes is pictured in this image.''
        \end{itemize}
    \end{minipage} 
    \begin{minipage}[c]{0.3\hsize}\centering
        \centering
        \includegraphics[width=1.75in]{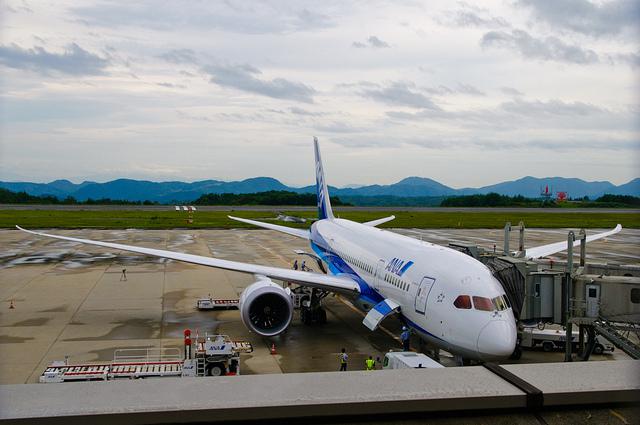}
        \caption*{}
    \end{minipage}
    \begin{minipage}[c]{0.75\hsize}\centering
        \begin{itemize}
            \setlength\itemsep{0.1em}
            \vspace{-2em}
            \item[-] ``An All Nippon Airways 777 sitting at a gate on the tarmac.''
            \item[-] ``a large air plane on a run way''
            \item[-] ``A jumbo jet being serviced at an airport.''
            \item[-] ``A large blue and white jetliner sitting on top of a tarmac.'', 
            \item[-] ``A large airliner preparing for departure at an airport.''
        \end{itemize}
    \end{minipage}
    \begin{minipage}[c]{0.3\hsize}\centering
        \centering
        \includegraphics[height=1.4in]{figures/dataset_examples/coco_27396}
        \caption*{}
    \end{minipage}
    \begin{minipage}[c]{0.75\hsize}\centering
        \begin{itemize}
            \setlength\itemsep{0.1em}
            \vspace{-2em}
            \item[-] ``A man jumping a horse over an obstacle.''
            \item[-] ``A person jumping a horse over an object.''
            \item[-] ``An equestrian competitor and his horse jumping over a stile''
            \item[-] ``A horse and jockey jump over bush hurdles .'', 
            \item[-] ``A rider and horse jump over a wooden brush obstacle.''
        \end{itemize}
    \end{minipage} \vspace{2em}
    \begin{minipage}[c]{0.3\hsize}\centering
        \centering
        \includegraphics[height=1in]{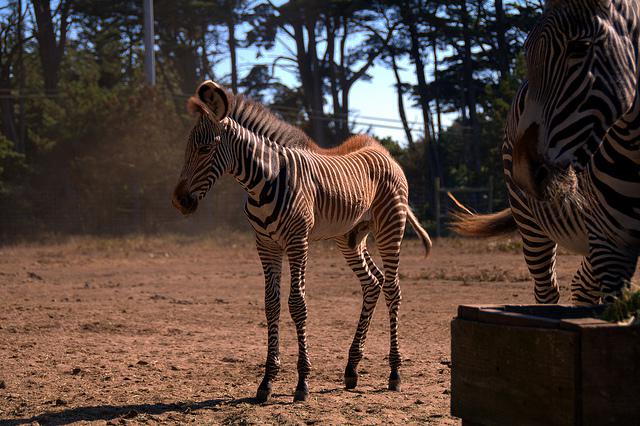}
        \caption*{}
    \end{minipage}
    \begin{minipage}[c]{0.75\hsize}\centering
        \begin{itemize}
            \setlength\itemsep{0.1em}
            \vspace{-2em}
            \item[-] ``A couple of zebra standing on top of a dirt field.''
            \item[-] ``Some zebras walking around in a field looking around''
            \item[-] ``Some very cute zebras in a big dusty field.''
            \item[-] ``A small zebra standing next to a bigger zebra.'', 
            \item[-] ``The baby Zebras stripes are much closer together than an adults.''
        \end{itemize}
    \end{minipage}
    \caption{Dataset examples: MS-COCO~\cite{lin2014microsofta}}
    \label{fig:app_datasets}
\end{figure}

\begin{figure}[!h]
    \ContinuedFloat
    \begin{minipage}[c]{0.35\hsize}\centering
        \centering
        \includegraphics[width=2in]{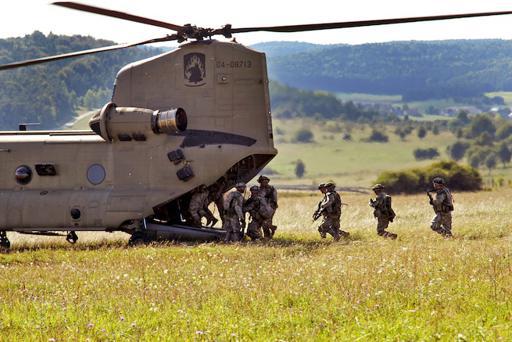}
        \caption*{}
    \end{minipage} \hfil
    \begin{minipage}[c]{0.64\hsize}\centering
        \begin{itemize}
            \item[] ``paratroopers load onto a helicopter.''
        \end{itemize}
    \end{minipage}
    \begin{minipage}[c]{0.35\hsize}\centering
        \centering
        \includegraphics[width=2in]{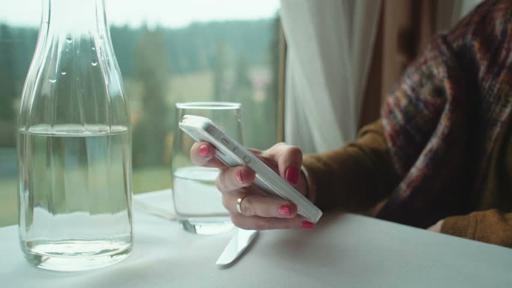}
        \caption*{}
    \end{minipage} \hfil
    \begin{minipage}[c]{0.64\hsize}\centering
        \begin{itemize}
            \item[] ``Close up hands of woman typing text message on smart phone in a cafe.''
        \end{itemize}
    \end{minipage}
    \begin{minipage}[c]{0.35\hsize}\centering
        \centering
        \includegraphics[width=2in]{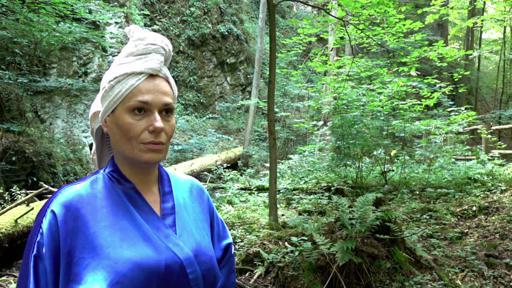}
        \caption*{}
    \end{minipage} \hfil
    \begin{minipage}[c]{0.64\hsize}\centering
        \begin{itemize}
            \item[] ``woman in a bathrobe is smiling to camera in the forest''
        \end{itemize}
    \end{minipage}
    \begin{minipage}[c]{0.35\hsize}\centering
        \centering
        \includegraphics[width=1.6in]{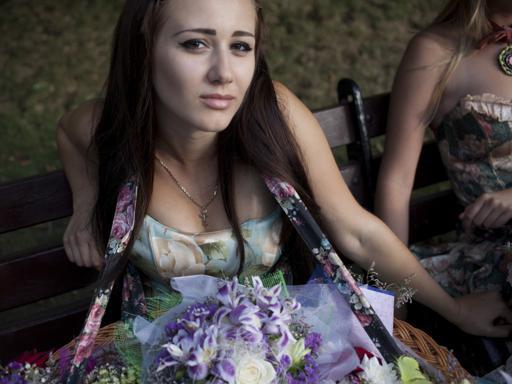}
        \caption*{}
    \end{minipage} \hfil
    \begin{minipage}[c]{0.64\hsize}\centering
        \begin{itemize}
            \item[] ``Girls in old time dresses selling flowers are pictured taking a rest of a bench.''
        \end{itemize}
    \end{minipage}
    \begin{minipage}[c]{0.35\hsize}\centering
        \centering
        \includegraphics[height=1.6in]{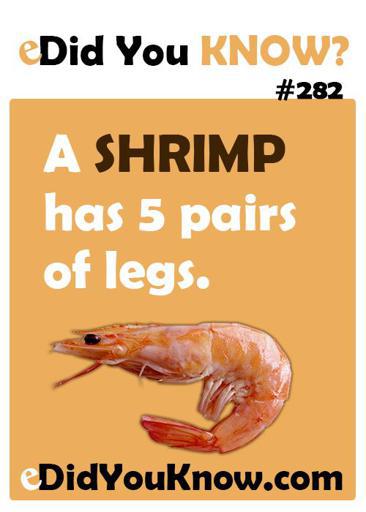}
        \caption*{}
    \end{minipage} \hfil
    \begin{minipage}[c]{0.64\hsize}\centering
        \begin{itemize}
            \item[] ``A shrimp has pairs of legs.''
        \end{itemize}
    \end{minipage}
    \caption{Dataset examples: Conceptual Captions~\cite{sharma2018conceptual}}
\end{figure}

\begin{figure}[!h]
    \ContinuedFloat
    \begin{minipage}[c]{0.35\hsize}\centering
        \centering
        \includegraphics[width=1.6in]{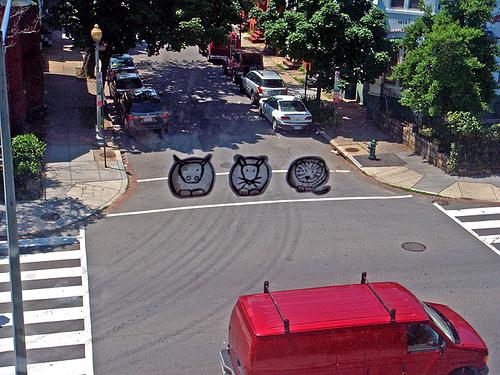}
        \caption*{}
    \end{minipage} \hfil
    \begin{minipage}[c]{0.64\hsize}\centering
        \begin{itemize}
            \vspace{-2em}
            \item[] ``Kenneth Phan \#7 A Day in the Life of DC is a photo project meant to capture a flavor of the region through the eyes of the participants. Participants submitted twelve photos taken on May 30, 2009. Photos by Kenneth Phan''
        \end{itemize}
    \end{minipage}
    \begin{minipage}[c]{0.35\hsize}\centering
        \centering
        \includegraphics[width=2in]{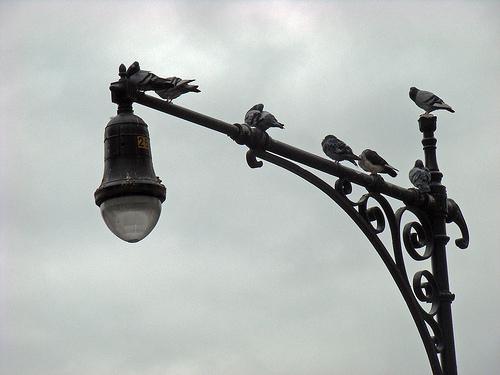}
        \caption*{}
    \end{minipage} \hfil
    \begin{minipage}[c]{0.64\hsize}\centering
        \begin{itemize}
            \vspace{-2em}
            \item[] ``Pombas New York - USA 27 de Setembro 2013''
        \end{itemize}
    \end{minipage}
    \begin{minipage}[c]{0.35\hsize}\centering
        \centering
        \includegraphics[width=1.6in]{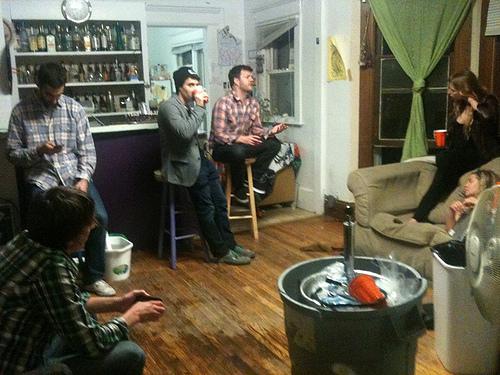}
        \caption*{}
    \end{minipage} \hfil
    \begin{minipage}[c]{0.64\hsize}\centering
        \begin{itemize}
            \vspace{-2em}
            \item[] ``Um, the girls that live at the house I lived in 13 years ago are huge @foursquare fans! \#amazing @ 600 euclid 4sq.com\/mPKNZv (posted via FlickSquare)''
        \end{itemize}
    \end{minipage}
    \begin{minipage}[c]{0.35\hsize}\centering
        \centering
        \includegraphics[height=1.5in]{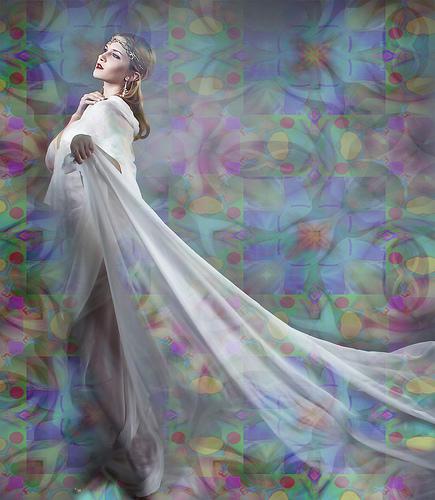}
        \caption*{}
    \end{minipage} \hfil
    \begin{minipage}[c]{0.64\hsize}\centering
        \begin{itemize}
            \vspace{-2em}
            \item[] ``squares Created for dA Users Gallery Challenge \#43 – Winter Stock 1 Model with thanks to Reine-Haru''
        \end{itemize}
    \end{minipage}
    \begin{minipage}[c]{0.35\hsize}\centering
        \centering
        \includegraphics[height=1.5in]{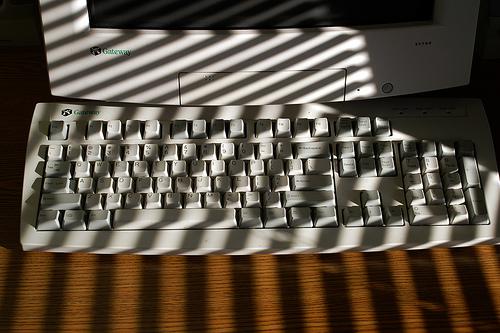}
        \caption*{}
    \end{minipage} \hfil
    \begin{minipage}[c]{0.64\hsize}\centering
        \begin{itemize}
            \vspace{-2em}
            \item[] ``Stripes and Squares Love the contrast of the light on the keyboard, stripes and squares''
        \end{itemize}
    \end{minipage}
    \caption{Dataset examples: YFCC~\cite{thomee2016yfcc100m}}
\end{figure}

%% file: figures/blip_examples.tex
\begin{figure}[!h]
    \begin{minipage}[c]{0.3\hsize}\centering
        \centering
        \includegraphics[height=1.3in]{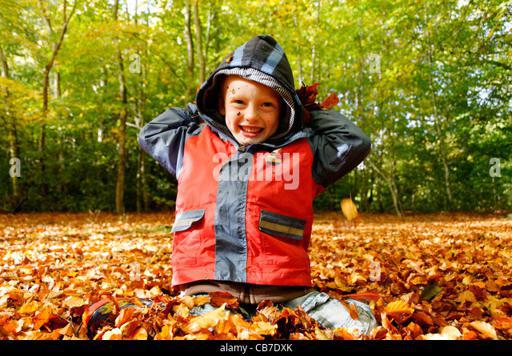} 
        \caption*{Dataset: CC}
    \end{minipage}
    \begin{minipage}[c]{0.77\hsize}\centering
        \begin{itemize}
            \setlength\itemsep{0.1em}
            \vspace{-2em}
            \item[-] ``portrait of a young boy sitting in the leaves in a park - stock image.''
            \item[-] ``toddler boy in a coat sitting on leaves with arms up to the air, smiling and laughing - stock photo.''
            \item[-] ``An image of a little boy sitting on the leaves in a park - stock image.''
        \end{itemize}
    \end{minipage}
    \begin{minipage}[c]{0.3\hsize}\centering
        \centering
        \includegraphics[height=1.3in]{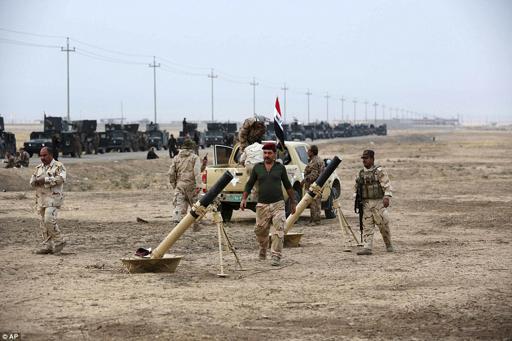} 
        \caption*{Dataset: CC}
    \end{minipage}
    \begin{minipage}[c]{0.77\hsize}\centering
        \begin{itemize}
            \setlength\itemsep{0.1em}
            \vspace{-2em}
            \item[-] ``The men are walking on a dirt ground with equipment in the background.''
            \item[-] ``military soldiers in uniforms carrying weapons and soldiers on their back in a desert''
            \item[-] ``Officials and soldiers stand in the desert, looking at a vehicle with missiles.''
        \end{itemize}
    \end{minipage}
    \begin{minipage}[c]{0.3\hsize}\centering
        \centering
        \includegraphics[height=1.3in]{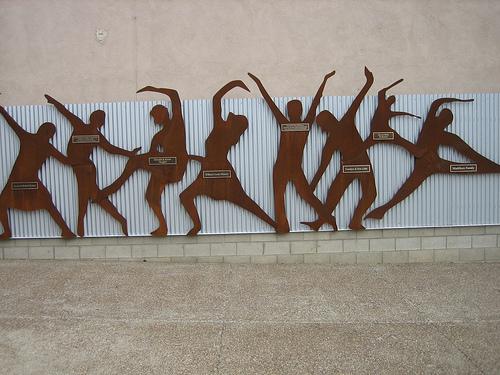} 
        \caption*{Dataset: YFCC}
    \end{minipage}
    \begin{minipage}[c]{0.77\hsize}\centering
        \begin{itemize}
            \setlength\itemsep{0.1em}
            \vspace{-2em}
            \item[-] ``Signs of various silhouettes of people dancing, standing, and laying on the street.''
            \item[-] ``lot of bronze colored women holding their arms up with their hands together in front of a metal wall art''
            \item[-] ``Sculptures on a wall of various silhouettes and dance positions.''
        \end{itemize}
    \end{minipage}
    \begin{minipage}[c]{0.3\hsize}\centering
        \centering
        \includegraphics[height=1.3in]{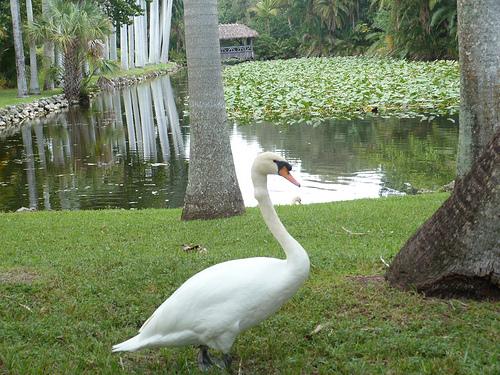} 
        \caption*{Dataset: YFCC}
    \end{minipage}
    \begin{minipage}[c]{0.77\hsize}\centering
        \begin{itemize}
            \setlength\itemsep{0.1em}
            \vspace{-2em}
            \item[-] ``I love the white swan in the foreground with the water behind him.''
            \item[-] ``an image group of white birds in a green area near some water''
            \item[-] ``the swan is standing on the green grass near the water''
        \end{itemize}
    \end{minipage}
    \caption{Random images from CC and YFCC alongside BLIP captions.}
    \label{fig:app_blip_examples}
\end{figure}

%% file: figures/synth_examples.tex
\begin{figure}[!h]
    \begin{minipage}[c]{0.32\hsize}\centering
        \centering
        \includegraphics[height=1.8in]{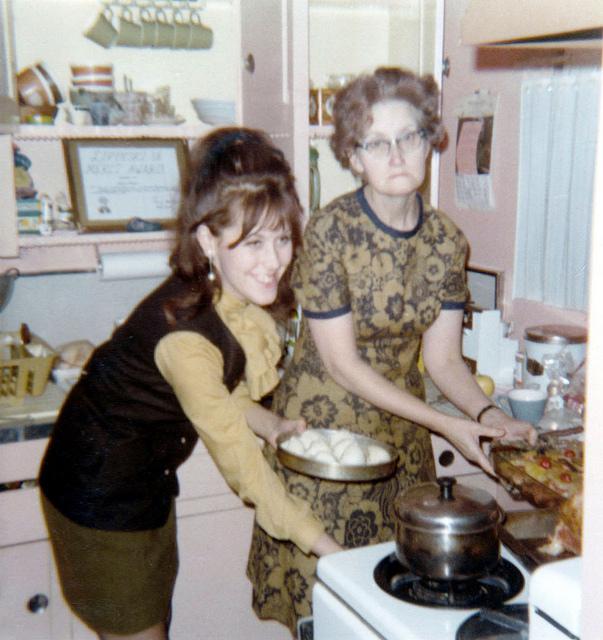}
    \end{minipage}
    \begin{minipage}[c]{0.74\hsize}\centering
        \emph{Complete and Consistent:}
        \begin{itemize}
            \setlength\itemsep{0.1em}
            \item[-] ``A photo of four bowls, a oven, seven cups, a refrigerator, two persons, a spoon, two cakes''
        \end{itemize}
        \emph{Incomplete and consistent:}
        \begin{itemize}
            \item[-] ``A photo of a person, six cups, three bowls, two cakes, a oven.''
        \end{itemize}
        \emph{Incomplete and Inconsistent:}
        \begin{itemize}
            \item[-] ``A photo of a kitchen, two women, two shot glasses.''
            \item[-] ``I see a oven, a kitchen, two mugs, a kitchen, a man.''
        \end{itemize}
        \caption*{}
    \end{minipage}
    \begin{minipage}[c]{0.32\hsize}\centering
        \centering
        \includegraphics[height=1.5in]{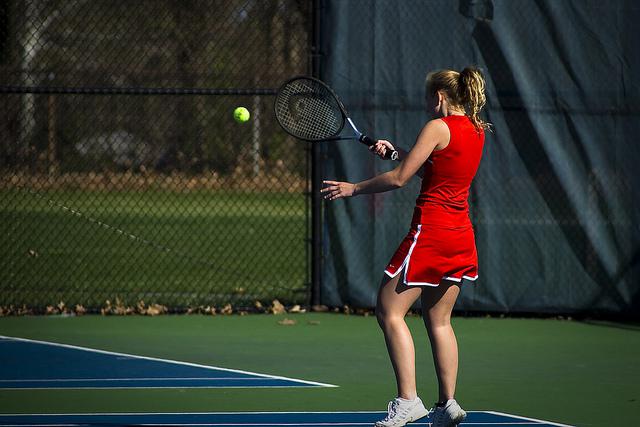}
    \end{minipage}
    \begin{minipage}[c]{0.74\hsize}\centering
        \emph{Complete and Consistent:}
        \begin{itemize}
            \item[-]``A photo of a person, a tennis racket, a sports ball, a car.''
        \end{itemize}
        \emph{Incomplete and consistent:}
        \begin{itemize}
            \item[-] ``A photo of a car, a person, a tennis racket.''
        \end{itemize}
        \emph{Incomplete and Inconsistent:}
        \begin{itemize}
            \item[-] ``a sports ball, a motorcar together.''
            \item[-] ``There is a woman.''
        \end{itemize}
    \end{minipage}
    \caption{Random image samples from MS-COCO alongside our synthetic captions.}
    \label{fig:app_synthetic_examples}
\end{figure}

%% file: figures/filt_examples.tex
\begin{figure}[!h]
    \begin{minipage}[c]{0.35\hsize}\centering
        \centering
        \includegraphics[width=1.6in]{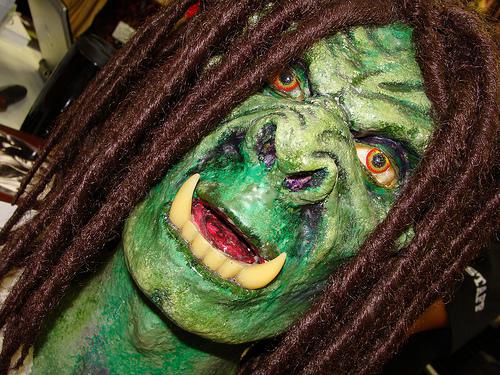}
        \caption*{}
    \end{minipage} \hfil
    \begin{minipage}[c]{0.64\hsize}\centering
        \begin{itemize}
            \item[] ``Orc/Troll There's a face only a mother could love.''
        \end{itemize}
    \end{minipage}
    \begin{minipage}[c]{0.35\hsize}\centering
        \centering
        \includegraphics[height=1.2in]{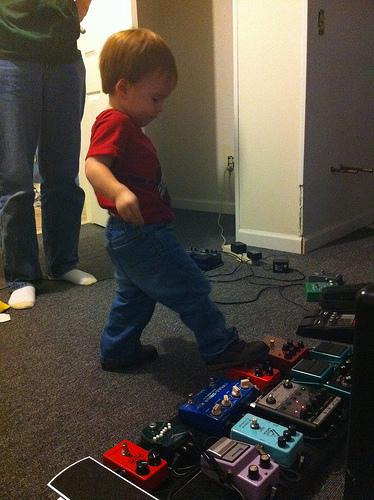}
        \caption*{}
    \end{minipage} \hfil
    \begin{minipage}[c]{0.64\hsize}\centering
        \begin{itemize}
            \item[] ``Pedal Board 9 Back Camera.''
        \end{itemize}
    \end{minipage}
    \begin{minipage}[c]{0.35\hsize}\centering
        \centering
        \includegraphics[height=1.2in]{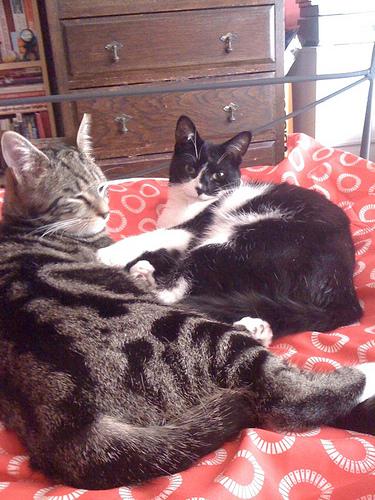}
        \caption*{}
    \end{minipage} \hfil
    \begin{minipage}[c]{0.64\hsize}\centering
        \begin{itemize}
            \item[] ``Kittens Morrissey and Marr relax on the bed.''
        \end{itemize}
    \end{minipage}
    \begin{minipage}[c]{0.35\hsize}\centering
        \centering
        \includegraphics[width=1.6in]{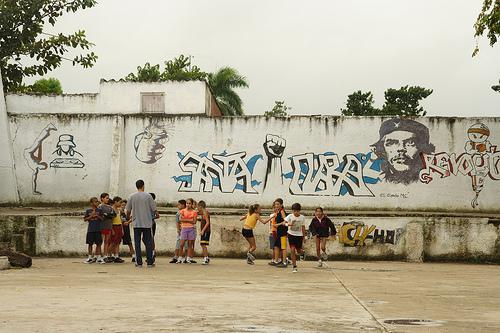}
        \caption*{}
    \end{minipage} \hfil
    \begin{minipage}[c]{0.64\hsize}\centering
        \begin{itemize}
            \item[] ``Park in Santa Clara Complete with Che graffiti.''
        \end{itemize}
    \end{minipage}
    \caption{Filtered image-caption pairs from the YFCC dataset (see Section~\ref{sec:prescription}).}
    \label{fig:app_fasttext_examples}
\end{figure}

%% file: figures/gpt_examples.tex
\begin{figure}[!h]
    \begin{minipage}[c]{0.85\hsize}
        \textbf{CC caption:} ``young beautiful businesswoman holding the documents and crossing the road in financial district .'' \\
        \textbf{GPT-J paraphrases:}
        \begin{itemize}
            \item[-] ``A women in business attire crossing the road in financial district.''
            \item[-] ``A young beautiful businesswoman is crossing the road.''
            \item[-] ``A young woman crossing the road in financial district.''
        \end{itemize}
        \caption*{}
    \end{minipage}
    \begin{minipage}[c]{0.85\hsize}
        \textbf{CC caption:} ``Island with waterfalls and rainbows floating above clouds in a dreamy sky -- stock photo \#'' \\
        \textbf{GPT-J paraphrases:}
        \begin{itemize}
            \item[-] ``An island with waterfalls and rainbows floating above clouds in a dreamy sky.''
            \item[-] ``A rainbow in the sky with a cloud and a rainbow in the water''
            \item[-] ``A rainbow floating in a dreamy sky and a waterfall''
        \end{itemize}
        \caption*{}
    \end{minipage}
    \begin{minipage}[c]{0.85\hsize}
        \textbf{CC caption:} ``Young friends eating pizza in the park'' \\
        \textbf{GPT-J paraphrases:}
        \begin{itemize}
            \item[-] ``The young friends are eating pizza in the park''
            \item[-] ``A group of young people eating pizza in the park''
            \item[-] ``Some kids are eating pizza in the park''
        \end{itemize}
        \caption*{}
    \end{minipage}
    \begin{minipage}[c]{0.85\hsize}
        \textbf{CC caption:} ``Actors attend the spring fashion show during fashion week.'' \\
        \textbf{GPT-J paraphrases:}
        \begin{itemize}
            \item[-] ``A fashion show that's attended by actors.''
            \item[-] ``People attending a spring fashion show during fashion week.''
            \item[-] ``The spring fashion show was held during fashion week.''
        \end{itemize}
        \caption*{}
    \end{minipage}
    \caption{Caption paraphrases generated using in-context learning with GPT-J.}
    \label{fig:app_gptj_examples}
\end{figure}

%% file: appendix/expts.tex
\section{Additional experiments}
\label{app:expts}

In Appendix Tables~\ref{tab:app_coco_comparison_basic}-\ref{tab:app_gpt}, we report per-task performance for all our experiments. 
In Appendix Table~\ref{tab:app_coco_comparison_basic}, we also illustrate the performance of SimCLR/CLIP models trained using the simpler data augmentations typically used for CLIP training (cf. Appendix~\ref{app:hyperparameters}).
One can see that both models perform worse with this modification---with the performance of CLIP dropping by 10\% and that of SimCLR by 50\%.

For COCO, we also consider a variant of SimCLR, which we refer to as SimCLR$_{+lab}$, that factors in label information in the transformation $T(x)$.
Specifically, for a given image $x$, $x_+$ is a data augmented version of another COCO image which has at least one object in common with $x$. 
We see that factoring label information does improve SimCLR's performance considerably, putting it between vanilla CLIP and \clipd{}.
However, note for typical pre-training datasets such as CC and YFCC, we do not have access to such ``expert'' object labels.
Instead, we can take advantage of captions to improve the equivalences learned by the model.

\input{tables/app_coco_clip_vs_simclr}
\input{tables/app_coco_div}

\input{tables/app_cc}
\input{tables/app_yfcc}

\input{tables/app_justify}
\input{tables/app_blip}
\input{tables/app_filter}
\input{tables/app_gpt}

%% file: tables/app_coco_clip_vs_simclr.tex

\begin{table}[!h]
	\centering
	\setlength{\tabcolsep}{2.5pt}
	\renewcommand{\arraystretch}{1.3}
	
\begin{tabular}{|l||c|c|c|c|c|c|c|c|}
\hline 
Model & SUP & SimCLR$_{-}$ & \simclr & SimCLR$_{+lab}$ & CLIP$_{-}$ & CLIP & \clipd \\
\hline
COCO & 90.5 $\pm$ 1.5 & 60.4 $\pm$ 2.4 & 88.9 $\pm$ 1.6 & 89.3 $\pm$ 1.5 & 84.9 $\pm$ 1.9 & 88.4 $\pm$ 1.7 & 89.8 $\pm$ 1.6 \\ \hline
Aircraft & 31.6 $\pm$ 0.9 & 2.3 $\pm$ 0.3 & 40.6 $\pm$ 1.0 & 47.0 $\pm$ 1.0 & 30.3 $\pm$ 1.0 & 41.4 $\pm$ 1.0 & 46.4 $\pm$ 1.0 \\
Birdsnap & 11.8 $\pm$ 0.4 & 0.7 $\pm$ 0.1 & 18.5 $\pm$ 0.5 & 20.8 $\pm$ 0.5 & 14.0 $\pm$ 0.4 & 17.6 $\pm$ 0.5 & 20.0 $\pm$ 0.5 \\
Cal101 & 65.8 $\pm$ 0.7 & 3.8 $\pm$ 0.3 & 71.5 $\pm$ 0.7 & 80.4 $\pm$ 0.6 & 53.6 $\pm$ 0.8 & 73.2 $\pm$ 0.7 & 78.4 $\pm$ 0.6 \\
Cal256 & 53.7 $\pm$ 0.5 & 3.1 $\pm$ 0.2 & 58.6 $\pm$ 0.4 & 65.7 $\pm$ 0.4 & 41.5 $\pm$ 0.5 & 60.4 $\pm$ 0.5 & 65.6 $\pm$ 0.5 \\
Cars & 21.7 $\pm$ 0.5 & 1.2 $\pm$ 0.1 & 31.4 $\pm$ 0.6 & 39.3 $\pm$ 0.7 & 23.4 $\pm$ 0.5 & 35.8 $\pm$ 0.6 & 41.5 $\pm$ 0.6 \\
CIFAR-10 & 74.8 $\pm$ 0.5 & 23.2 $\pm$ 0.5 & 82.1 $\pm$ 0.4 & 81.5 $\pm$ 0.5 & 74.0 $\pm$ 0.5 & 83.6 $\pm$ 0.4 & 84.6 $\pm$ 0.4 \\
CIFAR-100 & 46.7 $\pm$ 0.6 & 6.0 $\pm$ 0.3 & 57.3 $\pm$ 0.6 & 56.8 $\pm$ 0.6 & 50.4 $\pm$ 0.6 & 60.8 $\pm$ 0.6 & 62.5 $\pm$ 0.6 \\
DTD & 55.9 $\pm$ 1.4 & 6.2 $\pm$ 0.6 & 61.7 $\pm$ 1.3 & 60.3 $\pm$ 1.3 & 48.2 $\pm$ 1.4 & 65.7 $\pm$ 1.3 & 66.7 $\pm$ 1.3 \\
Flowers & 63.5 $\pm$ 0.7 & 4.6 $\pm$ 0.3 & 77.4 $\pm$ 0.6 & 81.4 $\pm$ 0.6 & 68.2 $\pm$ 0.7 & 80.5 $\pm$ 0.6 & 84.0 $\pm$ 0.6 \\
Food & 47.1 $\pm$ 0.4 & 4.0 $\pm$ 0.1 & 58.7 $\pm$ 0.3 & 56.4 $\pm$ 0.4 & 51.8 $\pm$ 0.4 & 60.9 $\pm$ 0.4 & 65.3 $\pm$ 0.4 \\
Pets & 45.9 $\pm$ 1.0 & 6.3 $\pm$ 0.5 & 57.3 $\pm$ 0.9 & 63.0 $\pm$ 0.9 & 44.6 $\pm$ 0.9 & 57.0 $\pm$ 0.9 & 61.2 $\pm$ 0.9 \\
SUN & 44.5 $\pm$ 0.4 & 1.3 $\pm$ 0.1 & 51.9 $\pm$ 0.4 & 52.2 $\pm$ 0.4 & 37.6 $\pm$ 0.4 & 50.8 $\pm$ 0.4 & 54.9 $\pm$ 0.4 \\ \hline
$\mu_{Tx}$ & 47.2 $\pm$ 0.2 & 5.2 $\pm$ 0.1 & 56.0 $\pm$ 0.2 & 58.7 $\pm$ 0.2 & 44.8 $\pm$ 0.2 & 57.5 $\pm$ 0.1 & 61.3 $\pm$ 0.2 \\
\hline
\end{tabular}
\caption{Extended comparison of transfer performance of supervised, SimCLR and CLIP pre-trained models. Here SimCLR$_{-}$ and  CLIP$_{-}$ denote models trained with the default CLIP data augmentation transforms instead of the SimCLR ones (cf. Appendix~\ref{app:hyperparameters}). SimCLR$_{+lab}$ refers to SimCLR models trained by picking $x_+$ to be a different image with the same label as $x$. \\}
\label{tab:app_coco_comparison_basic}
\end{table}

%% file: tables/app_coco_div.tex

\begin{table}[!t]
	\centering
	\setlength{\tabcolsep}{3pt}
	\renewcommand{\arraystretch}{1.4}
	
	\begin{tabular}{|l||c|c|c|c|c|}
		\hline 
		Model & CLIP & CLIP & \clipd & CLIP & \clipd \\
		Complete & \cmark & \xmark & \xmark & \xmark & \xmark \\
		Consistent & \cmark & \xmark & \xmark & \cmark & \cmark \\
		\hline
		COCO & 88.8 $\pm$ 1.7 & 88.4 $\pm$ 1.7 & 89.3 $\pm$ 1.6 & 88.3 $\pm$ 1.7 & 89.2 $\pm$ 1.5 \\ \hline
		Aircraft & 46.6 $\pm$ 1.0 & 44.5 $\pm$ 1.0 & 46.6 $\pm$ 1.0 & 45.6 $\pm$ 1.0 & 45.8 $\pm$ 1.0 \\
		Birdsnap & 18.9 $\pm$ 0.5 & 17.2 $\pm$ 0.5 & 18.6 $\pm$ 0.5 & 18.5 $\pm$ 0.5 & 19.1 $\pm$ 0.5 \\
		Cal101 & 77.3 $\pm$ 0.6 & 75.3 $\pm$ 0.7 & 76.8 $\pm$ 0.6 & 76.1 $\pm$ 0.7 & 76.0 $\pm$ 0.6 \\
		Cal256 & 63.3 $\pm$ 0.5 & 59.9 $\pm$ 0.5 & 63.0 $\pm$ 0.4 & 61.4 $\pm$ 0.5 & 63.6 $\pm$ 0.5 \\
		Cars & 42.4 $\pm$ 0.6 & 41.6 $\pm$ 0.6 & 42.7 $\pm$ 0.7 & 41.2 $\pm$ 0.6 & 42.8 $\pm$ 0.6 \\
		CIFAR-10 & 83.3 $\pm$ 0.4 & 82.4 $\pm$ 0.4 & 82.9 $\pm$ 0.4 & 83.7 $\pm$ 0.4 & 83.2 $\pm$ 0.4 \\
		CIFAR-100 & 60.5 $\pm$ 0.6 & 59.0 $\pm$ 0.6 & 58.9 $\pm$ 0.5 & 59.9 $\pm$ 0.5 & 60.1 $\pm$ 0.6 \\
		DTD & 64.3 $\pm$ 1.3 & 63.7 $\pm$ 1.3 & 66.1 $\pm$ 1.3 & 63.4 $\pm$ 1.2 & 65.2 $\pm$ 1.2 \\
		Flowers & 82.1 $\pm$ 0.5 & 78.3 $\pm$ 0.6 & 79.5 $\pm$ 0.6 & 79.3 $\pm$ 0.6 & 80.6 $\pm$ 0.6 \\
		Food & 61.4 $\pm$ 0.3 & 57.6 $\pm$ 0.4 & 60.9 $\pm$ 0.4 & 59.0 $\pm$ 0.3 & 61.9 $\pm$ 0.4 \\
		Pets & 60.0 $\pm$ 0.9 & 57.1 $\pm$ 1.0 & 58.8 $\pm$ 0.9 & 59.8 $\pm$ 0.9 & 60.6 $\pm$ 1.0 \\
		SUN & 52.1 $\pm$ 0.4 & 49.6 $\pm$ 0.4 & 53.1 $\pm$ 0.4 & 50.6 $\pm$ 0.4 & 52.7 $\pm$ 0.4 \\ \hline
		$\mu_{Tx}$ & 59.2 $\pm$ 0.1 & 56.6 $\pm$ 0.2 & 58.9 $\pm$ 0.2 & 57.7 $\pm$ 0.2 & 59.3 $\pm$ 0.2 \\
		\hline
		\end{tabular}
		\caption{The impact of intra-dataset variations in captions on CLIP's transfer performance. Here, we use synthetic captions for pre-training, constructed using COCO multi-object image labels. We vary whether these captions are consistent (i.e., do they use a single term to describe a given object?) and complete (i.e., do they describe all image objects?).
		We also consider a variant of CLIP, \clipd{}, which uses multiple captions per image. \\}
		\label{tab:app_coco_diversity}

\end{table}

%% file: tables/app_cc.tex

\begin{table}[!t]
	\centering
	\setlength{\tabcolsep}{2.5pt}
	\vspace{1em}
	\renewcommand{\arraystretch}{1.3}
	
	\begin{tabular}{|c||cccc|cccc|}
		\hline 
		Model & \multicolumn{4}{c}{\simclr{}} & \multicolumn{4}{|c|}{CLIP} \\ 
		Dataset size & 100K & 200K & 500K & 2M & 100K & 200K & 500K & 2M \\ \hline
		Aircraft & 40.5 $\pm$ 1.0 & 40.3 $\pm$ 1.0 & 39.3 $\pm$ 0.9 & 37.9 $\pm$ 1.0 & 35.5 $\pm$ 1.0 & 39.9 $\pm$ 1.0 & 41.6 $\pm$ 1.0 & 45.1 $\pm$ 1.0 \\
		Birdsnap & 20.2 $\pm$ 0.5 & 20.7 $\pm$ 0.5 & 20.6 $\pm$ 0.5 & 20.6 $\pm$ 0.5 & 15.1 $\pm$ 0.5 & 17.5 $\pm$ 0.5 & 19.8 $\pm$ 0.5 & 24.0 $\pm$ 0.6 \\
		Cal101 & 70.7 $\pm$ 0.7 & 70.3 $\pm$ 0.7 & 70.3 $\pm$ 0.7 & 69.0 $\pm$ 0.7 & 67.7 $\pm$ 0.8 & 73.5 $\pm$ 0.7 & 79.0 $\pm$ 0.6 & 84.8 $\pm$ 0.6 \\
		Cal256 & 57.7 $\pm$ 0.5 & 57.3 $\pm$ 0.5 & 57.3 $\pm$ 0.5 & 56.7 $\pm$ 0.5 & 54.4 $\pm$ 0.4 & 60.0 $\pm$ 0.5 & 65.9 $\pm$ 0.4 & 73.9 $\pm$ 0.4 \\
		Cars & 33.3 $\pm$ 0.6 & 31.2 $\pm$ 0.6 & 29.6 $\pm$ 0.6 & 27.5 $\pm$ 0.6 & 29.8 $\pm$ 0.6 & 33.8 $\pm$ 0.6 & 37.7 $\pm$ 0.6 & 42.6 $\pm$ 0.7 \\
		CIFAR-10 & 81.0 $\pm$ 0.5 & 80.4 $\pm$ 0.5 & 79.3 $\pm$ 0.5 & 79.8 $\pm$ 0.5 & 82.5 $\pm$ 0.4 & 83.9 $\pm$ 0.4 & 85.6 $\pm$ 0.4 & 86.8 $\pm$ 0.4 \\
		CIFAR-100 & 58.1 $\pm$ 0.6 & 57.4 $\pm$ 0.6 & 56.4 $\pm$ 0.5 & 56.4 $\pm$ 0.6 & 59.7 $\pm$ 0.6 & 63.2 $\pm$ 0.5 & 64.8 $\pm$ 0.5 & 67.8 $\pm$ 0.6 \\
		DTD & 62.8 $\pm$ 1.3 & 63.9 $\pm$ 1.2 & 64.5 $\pm$ 1.2 & 64.3 $\pm$ 1.3 & 63.7 $\pm$ 1.3 & 67.6 $\pm$ 1.3 & 70.3 $\pm$ 1.3 & 74.7 $\pm$ 1.2 \\
		Flowers & 80.8 $\pm$ 0.6 & 80.3 $\pm$ 0.6 & 80.2 $\pm$ 0.6 & 79.4 $\pm$ 0.6 & 76.5 $\pm$ 0.6 & 80.8 $\pm$ 0.6 & 85.0 $\pm$ 0.5 & 88.8 $\pm$ 0.5 \\
		Food & 57.6 $\pm$ 0.3 & 58.3 $\pm$ 0.4 & 57.0 $\pm$ 0.4 & 56.7 $\pm$ 0.4 & 56.6 $\pm$ 0.4 & 59.4 $\pm$ 0.4 & 62.7 $\pm$ 0.3 & 68.1 $\pm$ 0.3 \\
		Pets & 58.2 $\pm$ 0.9 & 57.9 $\pm$ 1.0 & 56.8 $\pm$ 0.9 & 55.9 $\pm$ 0.9 & 49.7 $\pm$ 1.0 & 53.5 $\pm$ 0.9 & 60.2 $\pm$ 0.9 & 65.2 $\pm$ 0.9 \\
		SUN & 49.4 $\pm$ 0.4 & 49.8 $\pm$ 0.4 & 49.7 $\pm$ 0.4 & 49.6 $\pm$ 0.4 & 45.9 $\pm$ 0.4 & 50.9 $\pm$ 0.4 & 55.3 $\pm$ 0.4 & 61.8 $\pm$ 0.4 \\
		\hline
		$\mu_{Tx}$ & 55.9 $\pm$ 0.2 & 55.3 $\pm$ 0.2 & 55.1 $\pm$ 0.2 & 54.5 $\pm$ 0.2 & 53.1 $\pm$ 0.2 & 57.0 $\pm$ 0.2 & 60.7 $\pm$ 0.2 & 65.3 $\pm$ 0.2 \\
		\hline
		\end{tabular}
		\caption{Transfer performance of SimCLR and CLIP models after pre-training on CC subsets. }
		\label{tab:app_cc}		

\end{table}

%% file: tables/app_yfcc.tex

\begin{table}[!t]
	\centering
	\setlength{\tabcolsep}{2.5pt}
	\renewcommand{\arraystretch}{1.3}
	\vspace{1em}
	
\begin{tabular}{|c||cccc|cccc|}
\hline 
Model & \multicolumn{4}{c}{\simclr{}} & \multicolumn{4}{|c|}{CLIP} \\ 
Dataset size & 100K & 200K & 500K & 2M & 100K & 200K & 500K & 2M \\ \hline
Aircraft & 39.5 $\pm$ 0.9 & 39.3 $\pm$ 1.0 & 38.0 $\pm$ 0.9 & 36.3 $\pm$ 0.9 & 17.0 $\pm$ 0.7 & 21.2 $\pm$ 0.8 & 41.5 $\pm$ 0.9 & 43.0 $\pm$ 0.9 \\
Birdsnap & 19.2 $\pm$ 0.5 & 18.9 $\pm$ 0.5 & 19.7 $\pm$ 0.5 & 19.0 $\pm$ 0.5 & 8.3 $\pm$ 0.4 & 10.4 $\pm$ 0.4 & 19.8 $\pm$ 0.5 & 26.2 $\pm$ 0.6 \\
Cal101 & 71.0 $\pm$ 0.7 & 71.1 $\pm$ 0.7 & 70.3 $\pm$ 0.7 & 68.4 $\pm$ 0.7 & 42.7 $\pm$ 0.7 & 51.4 $\pm$ 0.7 & 75.2 $\pm$ 0.7 & 82.1 $\pm$ 0.6 \\
Cal256 & 56.9 $\pm$ 0.5 & 58.5 $\pm$ 0.5 & 58.6 $\pm$ 0.5 & 57.7 $\pm$ 0.5 & 32.9 $\pm$ 0.4 & 38.2 $\pm$ 0.5 & 62.4 $\pm$ 0.5 & 70.5 $\pm$ 0.4 \\
Cars & 33.1 $\pm$ 0.6 & 29.8 $\pm$ 0.6 & 28.1 $\pm$ 0.6 & 26.8 $\pm$ 0.6 & 11.8 $\pm$ 0.4 & 15.5 $\pm$ 0.5 & 36.1 $\pm$ 0.7 & 37.4 $\pm$ 0.6 \\
CIFAR-10 & 80.4 $\pm$ 0.5 & 80.6 $\pm$ 0.4 & 80.2 $\pm$ 0.5 & 79.7 $\pm$ 0.5 & 71.1 $\pm$ 0.5 & 72.9 $\pm$ 0.5 & 83.5 $\pm$ 0.4 & 86.0 $\pm$ 0.4 \\
CIFAR-100 & 56.8 $\pm$ 0.5 & 58.2 $\pm$ 0.5 & 56.8 $\pm$ 0.5 & 57.2 $\pm$ 0.6 & 46.6 $\pm$ 0.6 & 47.7 $\pm$ 0.6 & 62.3 $\pm$ 0.6 & 66.2 $\pm$ 0.5 \\
DTD & 64.8 $\pm$ 1.2 & 67.0 $\pm$ 1.2 & 67.3 $\pm$ 1.2 & 67.0 $\pm$ 1.3 & 41.9 $\pm$ 1.3 & 49.9 $\pm$ 1.3 & 69.1 $\pm$ 1.2 & 74.3 $\pm$ 1.1 \\
Flowers & 80.9 $\pm$ 0.6 & 81.2 $\pm$ 0.6 & 80.5 $\pm$ 0.6 & 80.5 $\pm$ 0.6 & 47.6 $\pm$ 0.7 & 54.9 $\pm$ 0.8 & 83.4 $\pm$ 0.5 & 89.4 $\pm$ 0.4 \\
Food & 57.4 $\pm$ 0.4 & 57.9 $\pm$ 0.4 & 56.9 $\pm$ 0.4 & 57.4 $\pm$ 0.4 & 36.6 $\pm$ 0.4 & 43.0 $\pm$ 0.3 & 61.7 $\pm$ 0.3 & 67.4 $\pm$ 0.4 \\
Pets & 54.8 $\pm$ 1.0 & 55.4 $\pm$ 1.0 & 55.6 $\pm$ 0.9 & 55.6 $\pm$ 0.9 & 30.4 $\pm$ 0.9 & 34.0 $\pm$ 0.9 & 55.7 $\pm$ 1.0 & 61.9 $\pm$ 0.9 \\
SUN & 51.4 $\pm$ 0.4 & 52.9 $\pm$ 0.4 & 53.1 $\pm$ 0.4 & 53.2 $\pm$ 0.4 & 29.2 $\pm$ 0.4 & 34.7 $\pm$ 0.4 & 54.6 $\pm$ 0.4 & 62.8 $\pm$ 0.4 \\ \hline
$\mu_{Tx}$ & 55.5 $\pm$ 0.2 & 55.9 $\pm$ 0.2 & 55.4 $\pm$ 0.2 & 54.9 $\pm$ 0.2 & 34.7 $\pm$ 0.2 & 39.5 $\pm$ 0.2 & 58.8 $\pm$ 0.2 & 63.9 $\pm$ 0.2 \\
\hline
\end{tabular}
\caption{Transfer performance of SimCLR and CLIP models after pre-training on YFCC subsets. \\}
\label{tab:app_yfcc}
\end{table}

%% file: tables/app_justify.tex

\begin{table}[!t]
	\centering
	\setlength{\tabcolsep}{2.5pt}
	\renewcommand{\arraystretch}{1.3}
	\vspace{1em}
	
\begin{tabular}{|lll|ccccccccccc|}
\hline 
 Method & Size & Epochs & \zot{Aircraft} & \zot{Birdsnap} 
 & 
 \zot{Ctech101} & \zot{Cars} & 
 \zot{CIFAR10} & \zot{CIFAR100 \hspace{0.1ex}} & 
 \zot{DTD} & 
 \zot{Flowers} & \zot{Food-101} & \zot{Pets} & 
 \zot{SUN937} \\ \hline
 BYOL (\cite{tian2021divide}) &  100M & 1000 &  47.5 & 31.3 & 84.0 &  44.3 &  85.0 & 63.9 & 75.2  &  93.4&  67.9 & 71.1 & 63.4 \\
MoCLR (\cite{tian2021divide}) &  100M & 1000 & 45.6 & 29.4 & 85.6 & 41.1 & 87.8 & 69.9   & 75.8 & 92.9  & 67.7 & 67.7 &  63.4 \\
CLIP & 2M & 100 & {43.0} & {26.2} & {82.1} & {37.4} & {86.0} & {66.2} & {74.3} & {89.4} & {67.4} & {61.9} & {62.8}  \\
\hline
\end{tabular}
\caption{Comparison of our results to \cite{tian2021divide}. }
\label{tab:app_justify}
\end{table}

%% file: tables/app_blip.tex
\begin{table}[!t]
	\centering
	\setlength{\tabcolsep}{4pt}
	\renewcommand{\arraystretch}{1.4}
	
	\begin{tabular}{|c||cc|cc|c|c|} \hline
		Model & \multicolumn{4}{c|}{CLIP} & \multicolumn{2}{c|}{\clipd} \\ \hline
		Dataset & \multicolumn{2}{c|}{CC} & \multicolumn{2}{c|}{YFCC} & CC &  YFCC \  \\
		Dataset size & 100K & 500K & 100K & 500K & 100K & 100K \\ \hline
		Aircraft & 35.5 $\pm$ 1.0 & 41.6 $\pm$ 1.0 & 35.4 $\pm$ 0.9 & 42.8 $\pm$ 0.9 & 40.1 $\pm$ 1.0 & 41.3 $\pm$ 0.9 \\ 
		Birdsnap & 15.1 $\pm$ 0.5 & 19.8 $\pm$ 0.5 & 15.9 $\pm$ 0.5 & 20.7 $\pm$ 0.6 & 17.8 $\pm$ 0.5 & 19.2 $\pm$ 0.5 \\ 
		Cal101 & 67.7 $\pm$ 0.8 & 79.1 $\pm$ 0.6 & 67.9 $\pm$ 0.7 & 79.7 $\pm$ 0.6 & 75.1 $\pm$ 0.6 & 75.8 $\pm$ 0.6 \\ 
		Cal256 & 54.4 $\pm$ 0.5 & 65.9 $\pm$ 0.5 & 55.8 $\pm$ 0.5 & 67.6 $\pm$ 0.4 & 61.8 $\pm$ 0.5 & 62.6 $\pm$ 0.5 \\
		Cars & 29.8 $\pm$ 0.6 & 37.7 $\pm$ 0.6 & 29.6 $\pm$ 0.6 & 37.8 $\pm$ 0.6 & 37.3 $\pm$ 0.6  & 38.1 $\pm$ 0.6 \\
		CIFAR-10 & 82.5 $\pm$ 0.5 & 85.6 $\pm$ 0.4 & 82.9 $\pm$ 0.4 & 85.6 $\pm$ 0.4 & 83.6 $\pm$ 0.4 & 82.7 $\pm$ 0.4  \\
		CIFAR-100 & 59.7 $\pm$ 0.6 & 64.8 $\pm$ 0.5 & 60.9 $\pm$ 0.6 & 65.2 $\pm$ 0.5 & 62.2 $\pm$ 0.6 & 60.9 $\pm$ 0.6\\
		DTD & 63.7 $\pm$ 1.2 & 70.2 $\pm$ 1.2 & 64.1 $\pm$ 1.3 & 71.0 $\pm$ 1.3 & 67.7 $\pm$ 1.3 & 68.7 $\pm$ 1.2 \\
		Flowers & 76.5 $\pm$ 0.6 & 85.0 $\pm$ 0.5 & 77.7 $\pm$ 0.6 & 86.2 $\pm$ 0.5  & 81.4 $\pm$ 0.6 & 83.7 $\pm$ 0.6 \\
		Food & 56.6 $\pm$ 0.4 & 62.7 $\pm$ 0.3 & 57.3 $\pm$ 0.4 & 64.0 $\pm$ 0.3 & 59.6 $\pm$ 0.4 & 61.3 $\pm$ 0.3 \\
		Pets & 49.7 $\pm$ 1.0 & 60.2 $\pm$ 0.9 & 49.6 $\pm$ 0.9 & 61.6 $\pm$ 0.9 & 54.7 $\pm$ 0.9 & 56.6 $\pm$ 0.9 \\
		SUN & 45.9 $\pm$ 0.4 & 55.3 $\pm$ 0.4 & 47.7 $\pm$ 0.4 & 57.1 $\pm$ 0.4 & 52.5 $\pm$ 0.4 & 54.9 $\pm$ 0.4  \\ \hline
		$\mu_{Tx}$ & 53.7 $\pm$ 0.2 & 60.7 $\pm$ 0.2 & 54.8 $\pm$ 0.2 & 61.8 $\pm$ 0.2  & 57.8 $\pm$ 0.2 & 58.8 $\pm$ 0.2 \\ \hline
		\end{tabular}
		
		\caption{Effect of using BLIP captions for CC/YFCC images in CLIP training.}
		\label{tab:app_blip}

\end{table}

%% file: tables/app_filter.tex
\begin{table}[!t]
	\centering
	\setlength{\tabcolsep}{4pt}
	\renewcommand{\arraystretch}{1.4}
	
	\begin{tabular}{|c|cc|}
		\hline
		Dataset & CC & YFCC \\
		Dataset size & 100K & 500K \\ \hline
		Aircraft & 37.0 $\pm$ 1.0 & 41.0 $\pm$ 1.0 \\
		Birdsnap & 15.5 $\pm$ 0.5 & 21.1 $\pm$ 0.5 \\
		Cal101 & 71.1 $\pm$ 0.7 & 78.2 $\pm$ 0.7 \\
		Cal256 & 55.9 $\pm$ 0.5 & 64.9 $\pm$ 0.4 \\
		Cars & 30.9 $\pm$ 0.6 & 35.2 $\pm$ 0.6 \\
		CIFAR-10 & 82.9 $\pm$ 0.5 & 85.1 $\pm$ 0.4 \\
		CIFAR-100 & 59.3 $\pm$ 0.6 & 63.4 $\pm$ 0.6 \\
		DTD & 63.8 $\pm$ 1.3 & 71.8 $\pm$ 1.2 \\
		Flowers & 76.3 $\pm$ 0.7 & 84.3 $\pm$ 0.5 \\
		Food & 57.4 $\pm$ 0.4 & 64.1 $\pm$ 0.3 \\
		Pets & 52.7 $\pm$ 0.9 & 59.3 $\pm$ 0.9 \\
		SUN & 47.4 $\pm$ 0.4 & 56.4 $\pm$ 0.4 \\ \hline
		$\mu_{Tx}$ & 54.2 $\pm$ 0.2 & 60.4 $\pm$ 0.2 \\ \hline
		\end{tabular}
	\caption{Effect of caption filtering on CLIP's transfer performance.}  
	\label{tab:app_filter}
\end{table}

%% file: tables/app_gpt.tex
\begin{table}[!t]
	\centering
	\setlength{\tabcolsep}{4pt}
	\renewcommand{\arraystretch}{1.4}
	
	\begin{tabular}{|c|cc|} \hline
		Dataset & CC & COCO \\
		Dataset size & 200K & 120K \\ \hline
		Aircraft & 41.9 $\pm$ 0.9 & 44.7 $\pm$ 1.0 \\
		Birdsnap & 18.8 $\pm$ 0.5 & 18.6 $\pm$ 0.5 \\
		Cal101 & 77.4 $\pm$ 0.7 & 75.9 $\pm$ 0.6 \\
		Cal256 & 63.5 $\pm$ 0.4 & 62.8 $\pm$ 0.4 \\
		Cars & 38.2 $\pm$ 0.6 & 40.8 $\pm$ 0.6 \\
		CIFAR-10 & 84.0 $\pm$ 0.4 & 84.1 $\pm$ 0.4 \\
		CIFAR-100 & 62.5 $\pm$ 0.6 & 61.3 $\pm$ 0.6 \\
		DTD & 68.1 $\pm$ 1.2 & 65.3 $\pm$ 1.3 \\
		Flowers & 82.4 $\pm$ 0.6 & 81.9 $\pm$ 0.6 \\
		Food & 60.4 $\pm$ 0.4 & 62.0 $\pm$ 0.4 \\
		Pets & 56.0 $\pm$ 1.0 & 59.6 $\pm$ 1.0 \\
		SUN & 53.1 $\pm$ 0.4 & 51.9 $\pm$ 0.4 \\ \hline
		$\mu_{Tx}$ & 58.8 $\pm$ 0.3 & 58.9 $\pm$ 0.3 \\ \hline
\end{tabular}
\caption{Training \clipd{} models using additional captions generated via GPT-J paraphrasing.}
\label{tab:app_gpt}

\end{table}